\documentclass[fleqn,10pt]{wlscirep}
\usepackage[mathletters]{ucs}
\usepackage[T1]{fontenc}
\usepackage{cite}
\usepackage{amsmath,amssymb,amsfonts}
\usepackage{algorithmic}
\usepackage{graphicx}
\usepackage{multirow}
\usepackage{textcomp}
\usepackage{xcolor}
\usepackage{tabularray}
\usepackage{tabularx}
\usepackage{fancyhdr}
\usepackage{gensymb}
\usepackage{enumitem}
\usepackage{lineno}
\usepackage{float} 
\usepackage{multirow}
\usepackage{longtable}
\usepackage{booktabs}
\usepackage{array}
\usepackage{amsmath, array,graphicx}
\usepackage{caption,subcaption}
\usepackage{nameref}
\usepackage{amsmath,amssymb}
\usepackage{gensymb}
\usepackage{tabularray}
\usepackage{adjustbox}
\usepackage{multirow}

\def\BibTeX{{\rm B\kern-.05em{\sc i\kern-.025em b}\kern-.08em
    T\kern-.1667em\lower.7ex\hbox{E}\kern-.125emX}}

\PassOptionsToPackage{unicode}{hyperref}

\newcommand{\MS}[1]{{\color{black}{#1}}}

\newcommand{\AK}[1]{{\color{black}{#1}}}

\title{Tomato Maturity Recognition with Convolutional Transformers}

\author[1,2]{Asim Khan}
\author[3]{Taimur Hassan}
\author[2,4]{Muhammad Shafay}
\author[2,4]{Israa Fahmy}
\author[2,4]{Naoufel Werghi}
\author[1,2]{Seneviratne Mudigansalage}
\author[1,2,*]{Irfan Hussain}

\affil[1]{Department of Mechanical Engineering, Khalifa University, UAE}
\affil[2]{Khalifa University Center for Robotics and Autonomous Systems (KUCARS), Khalifa University, UAE}
\affil[3]{Department of Electrical, Computer and Biomedical Engineering, Abu Dhabi University, UAE}
\affil[4]{Department of Electrical Engineering and Computer Science, Khalifa University, UAE}

\affil[*]{Corresponding Author: Irfan Hussain (Irfan.Hussain@ku.ac.ae)}

\begin{abstract}
Tomatoes are a major crop worldwide, and accurately classifying their maturity is important for many agricultural applications, such as harvesting, grading, and quality control. In this paper, the authors propose a novel method for tomato maturity classification using a convolutional transformer. The convolutional transformer is a hybrid architecture that combines the strengths of convolutional neural networks (CNNs) and transformers.
Additionally, this study introduces a new tomato dataset named KUTomaData, explicitly designed to train deep-learning models for tomato segmentation and classification. KUTomaData is a compilation of images sourced from a greenhouse in the UAE, with approximately 700 images available for training and testing. The dataset is prepared under various lighting conditions and viewing perspectives and employs different mobile camera sensors, distinguishing it from existing datasets.
The contributions of this paper are threefold: 
Firstly, the authors propose a novel method for tomato maturity classification using a modular convolutional transformer. 
Secondly, the authors introduce a new tomato image dataset that contains images of tomatoes at different maturity levels. 
Lastly, the authors show that the convolutional transformer outperforms state-of-the-art methods for tomato maturity classification.
The effectiveness of the proposed framework in handling cluttered and occluded tomato instances was evaluated using two additional public datasets, Laboro Tomato and Rob2Pheno Annotated Tomato, as benchmarks. The evaluation results across these three datasets demonstrate the exceptional performance of our proposed framework, surpassing the state-of-the-art by 58.14\%, 65.42\%, and 66.39\% in terms of mean average precision scores for KUTomaData, Laboro Tomato, and Rob2Pheno Annotated Tomato, respectively.
This work can potentially improve the efficiency and accuracy of tomato harvesting, grading, and quality control processes.
\end{abstract}

\DeclareUnicodeCharacter{2212}{-}
\DeclareMathSymbol{\Beta}{\mathalpha}{operators}{"42}

\begin{document}

\flushbottom
\maketitle

\thispagestyle{empty}

\section{Introduction} 

Plants play a pivotal role in meeting global food demands. Among the most widely consumed vegetables are tomatoes, with annual production surpassing 180 million tons for the past seven years \cite{quinet2019tomato}. Commercially, tomatoes are typically harvested during the mature ripening stage.
This practice is primarily due to their firmness, extended shelf-life, and the potential to turn red after being removed from the plant \cite{bapat2010ripening}.
The decision to harvest at this stage is primarily influenced by consumer preferences for fresh tomatoes, particularly their colour and texture \cite{oltman2014consumer} and the need to minimize potential damage during transportation and other supply chain-related activities.

In academic research, the role of technology in optimizing agricultural practices is highly emphasized. A particular area of interest for scholars lies in the detection and classification of crops, where deep learning and image-processing techniques are utilized. Furthermore, automation in agriculture can enhance the working conditions of farmers and agricultural workers, who often face musculoskeletal disorders. The introduction of robots for crop monitoring and harvesting has proven highly beneficial, leading to significant improvements in production profits. These benefits are realized by streamlining the harvesting process, enhancing crop quality and yield, and reducing labour costs. These advantages have spurred extensive research over the past few decades, particularly on robotic technology's improvements and potential applications in agriculture. Whether referred to as "precision agriculture" or "low-impact farming", this approach forms an integral part of a broader shift within the agricultural industry. Additionally, advancements in computer vision can significantly enhance the agricultural sector by increasing efficiency and accuracy in various tasks, such as crop assessment and harvesting.

Machine learning (ML) methodologies significantly automate processes such as categorising plant diseases, fruit maturity grading, and automated harvesting methods~\cite{sangbamrung2020novel , septiarini2020maturity}. ML tools aid in monitoring plant health and predicting potential abnormalities at early stages \cite{ref7a}.
Over the years, various ML models have been developed, including Artificial Neural Networks and  Support Vector Machines (SVM) \cite{huang2018applications}. 

With the advent of  Deep Learning (DL), several new models such as VGG \cite{Simonyan15}, R-FCN \cite{https://doi.org/10.48550/arxiv.1605.06409}, Faster R-CNN \cite{ren2015faster}, and SSD \cite{liu2016ssd}; have been introduced,  providing fundamental frameworks to perform object detection and recognition tasks. 
Some of these methodologies find application in agricultural automation systems, aiding in identifying and classifying crops and their diseases. Notably, the advent of DL has led to promising results and methods in the agricultural domain.
Advancements in deep learning have made it possible to employ Convolutional Neural Networks (CNNs) in tasks such as fruit classification and yield estimation. For instance, Faster R-CNN \cite{ren2015faster} has been utilized for apple detection \cite{fu2020faster}, and YOLO has been applied to detect mangoes \cite{shi2020attribution}.
Sun et al. \cite{sun2018detection} proposed an enhanced version of the Faster R-CNN model, which demonstrated improved performance in detecting and identifying various parts of tomatoes, achieving a mean average precision (mAP) score of 90.7\% for the recognition of tomato flowers, unripened tomatoes, and ripe tomatoes. The optimized model exhibited a noteworthy reduction of approximately 79\% in memory requirements, suggesting the use of memory optimization techniques, such as parameter reduction methods or model compression techniques.
In another study, Liu et al. \cite{liu2020tomato} proposed a novel tomato detection model based on YOLOv3 \cite{yolov3}. Their model, which utilized a new bounding mechanism instead of conventional rectangular bounding boxes, enhanced the F1 score by 65\%.
Zhifeng et al. \cite{xu2020light} improved the YOLOv3-tiny model for ripe tomato identification, which achieved a 12\% improvement over its conventional counterpart in terms of the F1 score.
While detection models can identify and localize fruit regions within candidate scans, they often struggle to capture the contours and shapes of the fruits accurately. 
Segmentation methods can address this limitation by providing detailed information about fruit shapes and sizes through pixel-wise mask output.
For instance, as demonstrated by Yu et al. \cite{yu2019fruit}, the Mask R-CNN model was employed to successfully identify ripe strawberries, particularly those difficult to distinguish due to overlapping.
Similarly, Kang et al. \cite{kang2020fruit} employed the Mobile-DasNet model combined with a segmentation network to identify fruits, achieving accuracies of 90\% and 82\% for the respective tasks.

Ripeness is a critical factor in the quality and marketability of tomatoes. Traditionally, ripeness is assessed by human inspectors, who visually examine the tomatoes for colour, firmness, and other characteristics. However, this manual process is time-consuming, labour-intensive, and subjective. Early studies used simple features, such as the average RGB value of a tomato image, to classify ripeness.

Targeted fruit harvesting refers to the selective picking of ripe fruits, a complex task due to the unpredictable nature of crops and outdoor conditions. A vital example of this complexity is seen with tomatoes. They are a staple food crop widely grown worldwide but present a unique segmentation challenge due to their occlusion with leaves and stems, making it difficult to determine their ripeness. This is the reason for creating a new dataset that helps resolve these issues and provides a better perspective of tomato segmentation in complex environments. 
Introducing the KUTomaDATA dataset, with approximately 700 images obtained from greenhouses in Al Ajban, Abu Dhabi, United Arab Emirates, the authors address the pressing need for a comprehensive and diverse collection of tomato images to tackle real-life challenges in tomato farming. One of the novel features of KUTomaDATA lies in its representation of three distinct types of tomatoes: green, half-ripe, and fully ripe. This division into ripening stages, comprising "Fully Ripened," "Half Ripened," and "Un-ripened" tomatoes, provides a more nuanced and comprehensive dataset for researchers and practitioners. This dataset offers a unique and valuable resource for the computer vision community. 

In this research, the authors present a novel framework for the real-time segmentation of tomatoes and determining their maturity levels under diverse lighting and occlusion conditions. Here, our primary objective is to automate the process of tomato harvesting, potentially resulting in enhanced efficiency and reduced agricultural expenses. In addition to improving the harvesting process, accurately assessing tomato ripeness at the pixel level could also have other benefits. For example, it may allow for more precise sorting and grading of tomatoes, resulting in higher-quality final products. This could be particularly important for producers who export their tomatoes to different markets, as quality standards vary widely among countries.
 
In summary, this research can potentially bring about a paradigm shift in the harvesting and grading of tomatoes, which could have profound implications for the agricultural sector. By enhancing productivity and implementing stringent quality control measures, farmers may have the opportunity to boost their profitability while satisfying the increasing market demand for premium, environmentally friendly agricultural products. The main contributions of this study are outlined below:

\begin{enumerate}
    
    \item  The proposed approach provides a modular feature extraction and decoding method that separates the segmentation architecture, commonly referred to as the "meta-architecture," as illustrated in Figure \ref{fig1:block}.
    
    \item  Introducing a new dataset known as KUTomaData, captured under various Lighting, Occlusion, and Ripeness conditions from indoor glasshouse farms. Hence, this dataset provides many challenges to solve, giving it an edge over the existing datasets available to the research community.
        
    \item The proposed model is constrained via the $L_t$ loss function, enabling it to extract tomato regions from candidate scans that depict various textural, contextual, and semantic differences. Moreover, the $L_t$ loss function also ensures that the proposed model, at the inference stage, can objectively recognize different maturity stages of the tomatoes, irrespective of the scan attributes, for their effective cultivation. 
    
    \item  The proposed trained model is highly versatile and can be integrated into a mobile robot system designed for greenhouse farming. This integration would enable the robot to accurately detect and identify the maturity level of tomatoes in real time, which could significantly improve the efficiency and productivity of the farming process. 
    
\end{enumerate}
 
The remainder of the paper is organized as follows: Section \ref{m} delivers an in-depth discussion of the proposed method. Section \ref{ds} explores datasets. Section \ref{e} offers insights into the experiments and experimental procedures utilized. Section \ref{r} covers the evaluation results, Section \ref{as} covers the ablation study, and Section \ref{d} delves into a detailed discussion of the proposed framework. Section \ref{l} lists some of the limitations. Finally, Section \ref{c} concludes the paper.

\section{Related Work} 
\label{rw}
In this section, the authors highlight recent advances in precision agriculture proposed to assist farmers in effectively increasing their crop production, with a particular emphasis on tomatoes \cite{hasan2019deep}. To effectively organize the existing literature, the authors have categorized the methods into two groups: one group focuses on employing conventional techniques to enhance existing agricultural workflows, while the other group leverages modern computer vision schemes to enhance agricultural growth in terms of productivity, disease detection, and monitoring in natural farm environments \cite{dhanya2022deep}

\subsection{Traditional Methods in Precision Agriculture:}

Tomatoes are widely grown crops that have been the focus of many agricultural studies. Traditional approaches to improving tomato harvesting encompass various methods and principles for better managing these fruits against pests and diseases. These methods ultimately enhance overall agricultural productivity. Moreover, the evolution of these methods over the years has refined the foundation of traditional agricultural practices. Some of the standard methods proposed to improve agricultural workflows include:

Crop rotation is a strategic agricultural practice that involves the sequential cultivation of different crops across multiple seasons. Its purpose is to mitigate the negative impact of pests and diseases that specifically target certain crops while simultaneously improving soil fertility and overall crop yield \cite{FRANCIS2005318}.
Intercropping is a farming technique that involves cultivating two or more crops together in the same field concurrently \cite{VLAICULESCU2022329}. This method optimizes land utilization, promotes biodiversity, reduces the incidence of pests and diseases, and enhances soil fertility through nutrient complementarity.
Conventional irrigation methods encompass various systems such as flood, furrow, and sprinkler irrigation. These systems ensure a regulated water supply to crops, facilitating their optimal growth and development \cite{mitchell1993flood}
Furthermore, traditional agricultural practices have heavily relied on applying organic fertilizers, including crop residues, compost, and manure, to enhance soil fertility and provide essential nutrients to plants. These natural fertilizers contribute to long-term soil health and foster sustainable agricultural practices \cite{su12124859}.
Mechanical tillage involves using ploughs, harrows, and other machinery to prepare the soil for planting \cite{reicosky2003advances}. It serves multiple purposes, such as weed control, improved seedbed conditions, and incorporated nutrients into the soil. However, it is essential to note that mechanical tillage can also result in soil erosion and degradation.
Conventional pest and disease management methods predominantly rely on chemical pesticides and fungicides to control insects, weeds, and plant diseases. These methods aim to safeguard crops from damage and promote optimal growth. However, concerns have been raised regarding their potential adverse impacts on the environment and human health \cite{strand2000some}.
Acknowledging the strengths and limitations of these conventional agricultural practices is crucial to exploring opportunities for improvement and advancement in the field.

\subsection{Modern Computer Vision Methods for Precision Agriculture:}

Deep learning methods have recently attracted a lot of interest and have been increasingly utilized for the precise identification of tomato diseases and growth monitoring. Similarly, CNNs have also been utilized for tomato fertilization and disease detection \cite{sladojevic2016deep}.
These methods, built upon neural networks, are used to analyze large-scale datasets and derive insightful patterns for the precise detection and monitoring of tomatoes.
Sherafati et al. \cite{sherafati2022tomatoscan} proposed a framework for assessing the ripeness of tomatoes from RGB images. Sladojevic et al. \cite{sladojevic2016deep} utilized transfer learning to detect and classify tomato diseases. They achieved accurate disease classification by fine-tuning a pre-trained CNN network using a tomato disease dataset.
A. Khan et al.~\cite{10.1371/journal.pone.0243243} proposed a DeepLens Classification and Detection Model (DCDM) to classify healthy and unhealthy fruit trees and vegetable plant leaves using self-collected data and PlantVillage dataset~\cite{xu2018plantvillage}. Their experiments achieved an impressive 98.78\% accuracy in real-time diagnosis of plant leaf diseases.
Zheng et al.~\cite{zheng2022research} presented a YOLOv4 \cite{bochkovskiy2020yolov4} detector to determine tomato ripeness.
In contrast, Xu et al.~\cite{xu2022visual} utilized Mask R-CNN \cite{he2018mask} to differentiate between tomato stems and fruit.
Rong et al. \cite{rong2021peduncle} presented a framework based on YOLACT++ \cite{bolya2019yolact} for tomato identification. However, this model could not determine the tomatoes' ripeness due to the limited capability of the YOLACT++ framework in capturing and analyzing colour and textural features indicative of tomato ripeness. The YOLACT++ model primarily focuses on instance segmentation and object detection tasks without incorporating specific features or mechanisms to assess the ripeness of the tomatoes. As a result, the model's performance in accurately determining the ripeness level of the tomatoes was not satisfactory.

Incorporating semantic or instance segmentation models in agriculture can revolutionise how crops are assessed and harvested. While segmentation tasks are intricate, they offer the ability to identify objects and extract their semantic information at the pixel level. Such capabilities have become increasingly important for robots used in crop harvesting, where the first step is to detect, classify, and segment crops using computer vision methods \cite{arad2020development, xiong2020autonomous}.
For example, Liu et al. \cite{liu2020identification} employed UNet \cite{ronneberger2015unet}  to extract maize tassel. The authors achieved a high accuracy of 98.10\% and demonstrated the potential of using semantic segmentation for plant phenotyping.

Moreover, various studies have shown that using transformer models, such as ViT \cite{vit}, has improved the recognition of crops \cite{dosovitskiy2020image}. 
Likewise, transformers-based detection models have shown promising results in leaf disease detection and assessing the appearance quality of crops such as strawberries \cite{wang2021swingd, zheng2022swin, guo2022cst}.
Chen et al. \cite{rs14225853} used a Swin transformer \cite{liu2021swin} for detecting and counting wine grape bunch clusters in a non-destructive and efficient manner. Remarkably, their proposed approach achieved high recognition accuracy even in partial occlusions and overlapping fruit clusters.
Utilizing advanced computer vision techniques in agriculture can significantly enhance the effectiveness and precision of crop assessment and harvesting, ultimately boosting productivity and sustainability within the industry \cite{javaid2022enhancing}.

\section{Methods}
\label{m}
The precise segmentation of tomato maturity levels is important in various agricultural applications, such as harvesting, grading, and quality control. To address this challenge, we propose a novel framework that leverages advanced techniques, including encoder and transformer blocks, to process input scans effectively. \AK{The transformer block within the proposed model is derived from ViT \cite{dosovitskiy2020image}, and CMSA is the same as the multi-headed self-attention block in ViT \cite{dosovitskiy2020image}. In contrast to standard ViT variants, our approach involves the utilization of three transformer encoders arranged in a cascaded manner. This enables the generation of attentional characteristics, which are subsequently combined with convolutional features to extract various development stages of tomatoes effectively.} 
\begin{figure}[t]
\centering
\includegraphics[width=1\linewidth]{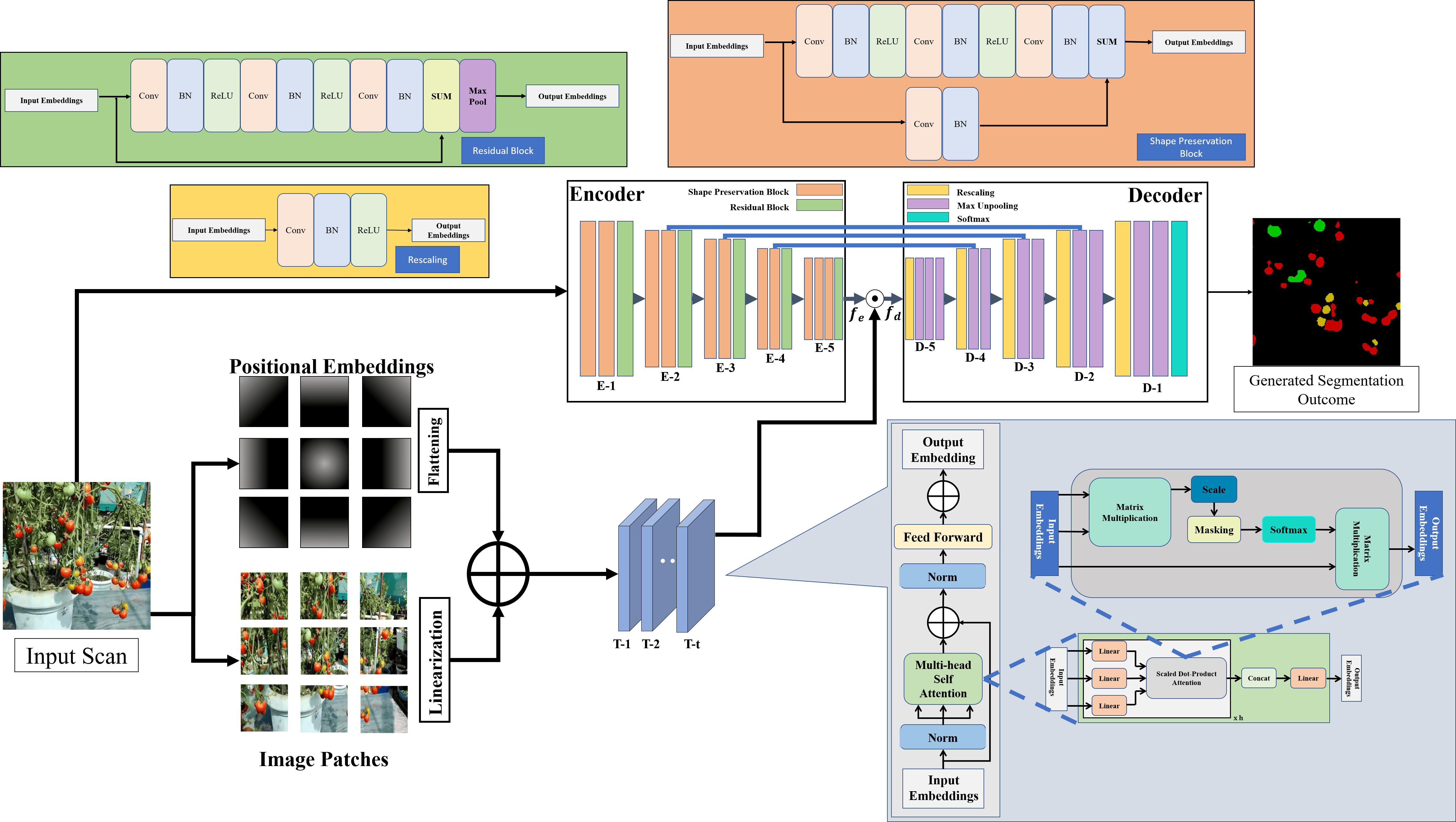}
\caption{\AK{An architectural diagram of the proposed framework for tomato maturity level recognition and grading. The proposed framework consists of the transformer, encoder, and decoder blocks. The input scan is initially passed to the transformer and encoder block. Across the transformer end, the input scan is divided into a set of image patches, against which the positional embeddings are computed. These positional embeddings and linear projections of the image patches are combined and are passed to the $t$-layered transformer block, which generates the projectional features to differentiate tomato grades. Similarly, the latent feature representations are computed from the input scan using the residual and shape preservation blocks at the encoder block. These latent space representations are then fused with the projectional features of the transformer end to boost the separation between different tomato grades. Finally, the decoder block removes extraneous elements through rescaling and max un-pooling operations, resulting in accurate segmentation and grading of tomato maturity levels.}}
\label{fig1:block}
\end{figure}
Initially, the input image is passed to the encoder and transformer blocks. The latent feature representations are computed from the input image using the residual and shape preservation blocks at the encoder block. Similarly, the input image is divided into $n$ number of image patches at the transformer end, against which $n$ positional embeddings are computed. These positional embeddings and linear projections of the image patches are combined and are passed to the $t$-layered transformer block to generate the projectional features via a contextual multi-head self-attention mechanism to differentiate between different tomato grades. We want to mention that the distinctive aspect of the transformer model (used in the proposed scheme) as compared to the conventional ViT is the number of stacked transformer blocks. \MS{In the original Vision Transformer (ViT) architecture, the ViT model consists of a stack of 8 identical transformer blocks. However, within the proposed scheme, we used only 3 stacked transformer blocks. The reason for using 3 stacked transformer blocks is because we achieved optimal trade-off between performance and computational complexity with this configuration toward recognizing different maturity stages of tomatoes. Adding more transformer blocks can increase further the performance of the proposed system but at the expense of adding excessive computational cost which we avoided by the current model design choice.} 
Finally, the decoder block removes extraneous elements through rescaling and max un-pooling operations, resulting in accurate segmentation and grading of tomato maturity levels. The subsequent sections provide a comprehensive overview of each block within the proposed framework:

\subsection{Transformer Block}

\AK{The proposed model incorporates a transformer block composed of \emph{t} encoders. Empirically, \emph{t} is set to 3, giving rise to encoders T-1, T-2, and T-3, which are cascaded together to generate \emph{$p_t$}.}
Initially, the input image \emph{x} is partitioned into non-overlapping, square-shaped patches denoted by {$x^{p}\epsilon{R}^P{^x}^{P}{^x}^{C_h}$}, where \emph{P} indicates the resolution of \emph{$x^p$} determined by the equation {$P=\sqrt{\frac{\text{RC}x}{\text{n}_p}}$}.
Here, \emph{$n_p$} represents the total number of patches. The positional embeddings \emph{$x^e_i$} corresponding to patch \emph{$x^p_i$} are then generated, i.e., {$x^{e}\epsilon{R}^P{^x}^{P}{^x}^{C_h}$}.
Subsequently, the flattened projections, i.e., \emph{$f_p(x^e_i)$}, are computed.
In a similar manner, the linear projection for patch \emph{$x^p_i$}, denoted as \emph{$l_t(x^p_i)$}, is obtained. Both \emph{$f_p(x^e_i)$} and \emph{$l_t(x^p_i)$} are resized to \emph{l} dimensions, and the sequenced embeddings for patch \emph{$x^p_i$} are computed by adding \emph{$l_t(x^p_i)$} to \emph{$f_p(x^e_i)$}, i.e., \emph{$q_i = l_t(x^p_i ) + f_p(x^e_i)$}.
By repeating this process for all the \emph{$n_p$} patches, the combined projections, \emph{$q^o$}, are generated, expressed as follows:
 \begin{equation}
     q^o=[l_t(x^{p}_0);(l_t(x^{p}_1)); . . .;l_t(x^p_{n{{_p}{_{}-1}}})] + [f_p(x^e_0); f_p(x^e_1); ... ; f_p(x^e_{n{{_p}{_{}-1}}})],
 \end{equation}

Or

\begin{equation}
{q^o = [q_0; q_1; ... ; q_{n{_p}{_{}-1}})]}:
 \end{equation}
This process allows the model to capture spatial information from the image and create a representation that the transformer block can further process.
The next step involves passing the combined projections $q^o$ to T-1, where each head j normalises $q^o{_j}$ to produce $q\acute{}_j{^o}$. 
Then, $q\acute{}_j{^o}$ is decomposed into a query ($Q_j$), key ($K_j$), and value ($V_j$) pairs using learn-able weights, with 
$Q_j = q\acute{}_j{^o}w_q$, 
$K = q\acute{}_j{^o}w_k$, and 
$V = q\acute{}_j{^o}w_v$. 
The contextual self-attention at head j (i.e., $A_j$) is then computed by combining $Q_j$ and $K_j$ through scaled dot product, and their resulting scores are merged with $V_j$. 

This computation is expressed below:

\begin{equation}
A_{j}(q\acute{}_j{^o}; Q_j,K_j,V_j )=\sigma(\frac{Q_jK{^T_j}}{\sqrt{l}})V_j,
 \end{equation}

The soft-max function $\sigma$ is applied element-wise to the output of the scaled dot product in each head. 
Furthermore, the contextual self-attention maps from all the leaders are concatenated to produce the contextual multi-head self-attention distribution $\varphi CMSA(\acute{q}^o)$, which is given by:

\begin{equation}
\varphi CMSA(\acute{q}^o) = [A_0(\acute{q}^o_j; Q_0; K_0; V_0); A_1(\acute{q}^o_j; Q_0; K_0;
V_0);.......; A_{h{-}1}(\acute{q}_j{^{h-1}}; Q_{h-1}; {K_{h-1}}; V_{h-1})]
 \end{equation}

This process enables the model to capture relationships and dependencies within the input patches. In addition to this, the contextual multi-head self-attention distribution $\varphi CMSA(\acute{q}^o)$ is combined with $q^{o}$, and the resulting embeddings are normalised and fed into the normalised feedforward block, which generates the T-1 latent projections $(p_{T1})$.

\begin{equation}
p_{T1} = \phi f((\varphi CMSA(\acute{q}^0) + q^o)\acute{} ) + (\varphi CMSA (\acute{q}^0) + qo)\acute{}
 \end{equation}
This process aims to generate more powerful and informative representations of the input data, which subsequent components in the model can further process.
After applying the learnable feed-forward function $\phi f(:)$, the resultant embeddings are normalised and passed through the normalised feedforward block to generate T-1 latent projections $(p_{T1})$. 
These projections are then passed to T-2, which produces $p_{T2}$ similarly. $p_{T2}$ is then passed to the T-3 encoder, which generates $p_{T3}$ projections. 
Here, $p_t = p_{T3}$. These projections are fused with $f_e$ to produce $f_d$. Finally, $f_d$ is passed to the decoder block to extract the instances of tomato objects.

\subsection{Encoder}
 
The encoder block in \emph{E} is responsible for creating the latent feature distribution \emph{$f_e(x)$} from the input tomato images $x\epsilon \mathbb{R}^{RxCXC_h}$, where \emph{R} represents rows, \emph{C} represents columns, and $C_h$ represents channels of \emph{x}. 
Unlike traditional pre-trained networks, \emph{E}'s encoder comprises five levels.
\emph{(E-1 to E-5)}, each with three to four shape preservation and residual blocks. 
These blocks empower the encoder to generate precise contextual and semantic representations of the targeted items during image decomposition while concurrently producing distinct feature maps. 
The encoder consists of 11 shape preservation blocks (SPBs) and five residual blocks (RBs), each with four convolutions, four batch normalisations (BNs), two ReLUs for SPBs, and three convolutions, three BNs, two ReLUs, and one max pooling for RBs.
The encoder's learned latent features \emph{($f_e$)}, after being fine-tuned, are effective in distinguishing the maturity level of one tomato from another. 
However, they may also produce false positives when differentiating between occluded regions of tomato objects, as their features are highly correlated. 
To mitigate this issue, the authors convolve \emph{$f_e$} with the transformer projections $p_t$ to enhance the distinction of inter-class distributions. 
\MS{The resulting fused feature representations \(f_d = f_e * p_t\) enhance similarities between \(f_e\) and \(p_t\), suppressing heterogeneous representations and significantly reducing false positives. \(f_e\) is convolved with \(p_t\) to produce \(f_d\), forwarded to the decoder. Convolution is a mathematical operation transforming one sequence using another, often termed an image, signal, or feature vector as the first input, and a filter as the second \cite{tensorflow_conv1d}. In the expression \(f_d = f_e * p_t\), \(p_t\) acts as a filter transforming the \(f_e\) feature vector to yield \(f_d\). These fused features then pass to the decoder, reconstructing the input image with segmented tomatoes.
}

\subsection{Decoder}

\noindent
The decoder block comprises several components that work together to segment tomato objects. It consists of 11 maximum unpooling layers, five rescaling layers, and a softmax layer. The unpooling layer plays a crucial role in recovering the spatial information lost during encoding. These layers help restore the original size and shape of the segmented objects. Each rescaling layer has a convolutional layer, batch normalization, and ReLU activation. Skip connections are also established between the encoder and decoder blocks to address the degradation problem that can occur during the segmentation of tomato objects. These connections enable the flow of information from earlier layers in the network to later layers. By doing so, the network can utilize low-level features from the encoder to refine and enhance the segmentation results in the decoder.
Following the successful segmentation process, a softmax layer is applied. This layer assigns each pixel in the segmented image to one of the tomato object categories based on its estimated maturity level. The softmax function computes the probability distribution over the categories, ensuring that each pixel is assigned to the most appropriate category. In conclusion, the proposed framework leverages the strengths of the encoder, transformer, and decoder blocks to achieve precise segmentation and grading of tomato maturity levels. The model efficiently collects spatial information, captures relationships among input patches, and enhances the differentiation between different tomato grades by utilizing learned latent features, contextual multi-head self-attention processes, and feature representation fusion. The decoder block refines the segmentation results and generates precise classifications with its unpooling layers, rescaling layers, and skip connections.


\subsection{Proposed $L_t$ Loss Function}

During the training phase, the model is constrained by the proposed loss function, referred to as $L_t$, which identifies and extracts tomato objects from input images.
The $L_t$ loss function comprises two components: $L_{s1}$ and $L_{s2}$.
By integrating these sub-objectives into the loss function, the model can be trained and subjected to a more extensive array of potential network defects.
This approach proves particularly useful when dealing with an imbalanced distribution of background and foreground pixels in the input scan, as it often leads to significantly smaller defect regions than the background region.
In such cases, $L_{s1}$ effectively minimises errors at the pixel level, enabling the model to perform segmentation tasks despite the imbalanced distribution of pixels.

However, attaining convergence through $L_{s1}$ presents challenges due to the possibility of the gradient of $L_{s1}$ to overshoot when the predicted logits and ground truths have smaller values.
To mitigate this issue, $L_{s2}$ is introduced into the $L_t$ loss function, allowing the model to converge even when dealing with smaller values of predicted logits and ground truths.
Moreover, the balance between $L_{s1}$ and $L_{s2}$ within $L_t$ is controlled by the hyperparameters $\beta_1$ and $\beta_2$. Mathematically, the objective functions can be expressed as follows:

\begin{equation}
    L_t = \beta_1 L_{s1} + \beta_2 L_{s2},
    \label{eq:eq1}
\end{equation}

\noindent where

\begin{equation} \label{eq:eq2}
    L_{s1} = \frac{1}{b_s} \sum_{i=0}^{b_s-1} \left(1 - \frac{2 \sum_{j=0}^{c_{se}-1} T^{se}_{i,j} p(\mathcal{L}_{i,j}^{se, \tau})} {\sum_{j=0}^{c_{se}-1} \left( (T^{se}_{i,j})^2 + p(\mathcal{L}^{se,\tau}_{i,j})^2 \right) }\right),
\end{equation}

\noindent and

\begin{equation} \label{eq:eq3}
L_{s2} = -\frac{1}{b_s}\sum\limits_{i=0}^{b_s-1}\sum\limits_{j=0}^{c_{se}-1} T_{i,j}^{se}\log(p(\mathcal{L}_{i,j}^{se, \tau})).
\end{equation}

The notation used in the context is as follows:
$T^{se}{i,j}$ denotes the ground truth label for the $i^{th}$ sample belonging to the $j^{th}$ tomato classes, namely full ripe, half ripe, and green.
$p(\mathcal{L}{i,j}^{se, \tau})$ indicates the predicted probability distribution obtained from the output logit $\mathcal{L}{i,j}^{se, \tau}$ for the $i^{th}$ sample and $j^{th}$ net defects category. This probability distribution is generated using the softmax function, and $\tau$ is a temperature constant used to soften the probabilities, ensuring robust learning of tomato classes.
$b_s$ signifies the batch size.
$c{se}$ represents the total number of classes, corresponding to the different tomato maturity levels considered.

\section{Datasets}
\label{ds}
This study leverages three different datasets, namely KUTomaData, Laboro Tomato \cite{Laboroai}, and Rob2Pheno \cite{10.3389/fpls.2020.571299}, to address various aspects of the research. Each dataset serves a specific role in contributing to the overall objectives of the study. Below, the authors provide detailed explanations for the characteristics and purposes of each dataset employed in this investigation.

\begin{figure*}[h!]
\centering
\includegraphics[width=0.65\textwidth]{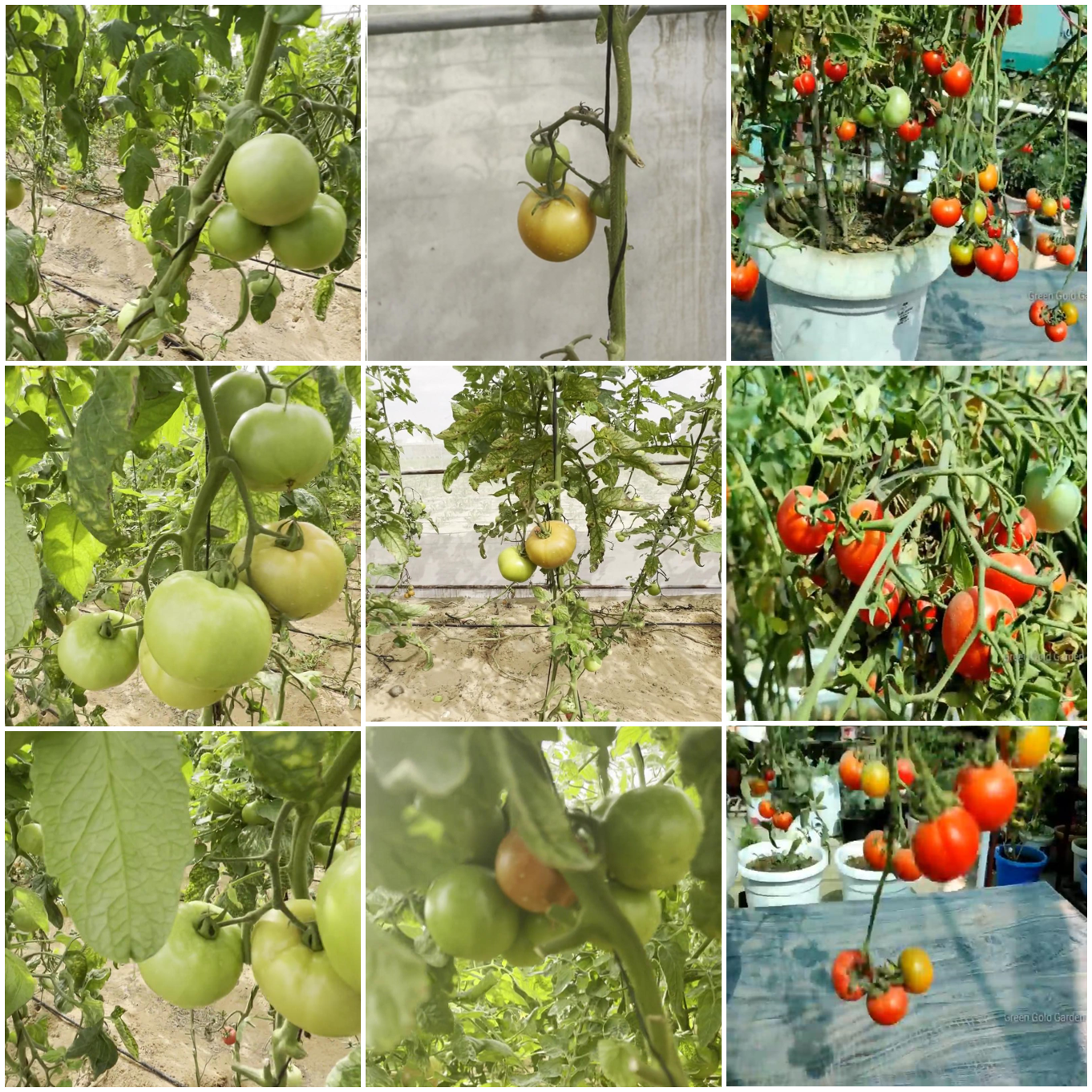}
 \caption{The dataset of tomato images contains samples of tomatoes captured in different stages of ripeness and under varying lighting conditions and occlusion. The images in the dataset are organized into three columns. The first column showcases unripened tomatoes, the second column shows half-ripe and unripened tomatoes, and the third column presents fully-ripened tomatoes with some half-ripened and some unripened tomatoes. This division allows for clear differentiation and visual representation of the different ripeness stages of the tomatoes in the dataset.}
\label{fig:fig2}
\end{figure*}

\subsection{KUTomaData:}

This dataset was collected from greenhouses in Al Ajban, Abu Dhabi, United Arab Emirates, and we have named it KUTomaData.
This dataset consists of approximately 7

00 images. The participants used mobile phone cameras to capture imagery from these greenhouses.
The dataset encompasses three distinct types of tomatoes: green, half-ripe, and fully ripe.
The ripening stages are classified into three categories:

\textbf{Fully Ripened:} This category represents tomatoes that have reached their optimal ripeness and are ready to be harvested. They exhibit a uniform red colouration, with at least 90\% of the tomato's surface filled with red colour.

\textbf{Half Ripened:} Tomatoes in this category are in a transitional ripening stage. They appear greenish and require more time to ripen fully. Typically, these tomatoes are red on 30\% to 89\% of their surface.

\textbf{Un-ripened:} This category encompasses tomatoes in the early ripening stages. They are predominantly green or white, with occasional small patches of red. These tomatoes have less than 30\% of their surface filled with red colour.

\AK{
The authors included images with varying hues, textures, and occlusion backdrops to ensure the dataset accurately mirrored real-world conditions.
The complexity of the dataset is heightened by the diverse backgrounds of the images, which exhibit varying densities and hues of tomatoes and leaves. This variability in the background composition adds intricacy to the dataset, making it more challenging and representative of real-world scenarios. The other challenging factors, such as complex environments, different lighting conditions, occlusion, and variations in tomato maturity levels and densities, were deliberately incorporated to ensure that the dataset accurately represents most real-world situations.}

The images presented in Figure \ref{fig:fig2} provide a visual presentation of the complexity of the dataset, with intricate backdrops for each tomato category and diverse illuminations and stages in most images.
This comprehensive and challenging dataset is suitable for training and testing the model's performance under realistic conditions.

\subsection{Laboro Tomato: Instance Segmentation \cite{Laboroai}: }

The Laboro Tomato dataset is a valuable collection of images that provides an in-depth exploration of the growth stages of tomatoes as they undergo the ripening process. With a total of 1,005 images, the dataset comprises 743 images for training and 262 images for testing. The dataset is curated to cater specifically to object detection and instance segmentation tasks, making it highly suitable for our research area. One notable aspect of the Laboro Tomato dataset is the inclusion of two distinct subsets of tomatoes, which are categorized based on size. This categorization adds an additional dimension to the dataset, allowing researchers to investigate the impact of tomato size on the performance of object detection and instance segmentation models.

To ensure the dataset's diversity and real-world relevance, the images were captured using two separate cameras, each with its unique resolution and image quality. The usage of different cameras introduces variations in image characteristics, such as colour rendition and sharpness, which can challenge the performance of computer vision models and better simulate real-world scenarios.

\subsection{Rob2Pheno Annotated Tomato \cite{10.3389/fpls.2020.571299}:}

Afonso et al. \cite{10.3389/fpls.2020.571299} conducted a research study focused on tomato fruit detection and counting in greenhouses using deep learning techniques. For this purpose, they utilized the Rob2Pheno Tomato dataset, which comprises RGB-D images of tomato plants captured in a production greenhouse setting. The images in this dataset were acquired using Real-sense cameras, which can capture both colour information and depth data. This additional depth information offers a three-dimensional perspective of the scene, providing valuable spatial context to the dataset.

Moreover, the Rob2Pheno Tomato dataset includes object instance-level ground truth annotations of the fruit. These annotations precisely identify the location and boundaries of individual tomato fruits within the images.
Regarding data volume, the dataset consists of 710 images for training purposes and 284 images for testing purposes. Data augmentation methods were applied during the training phase to enhance the dataset's diversity and improve the generalization ability of the models.


This paper presents a novel segmentation approach to extract and grade tomato maturity levels using RGB images acquired under various lighting and occlusion conditions.
Upon understanding the textures of the tomato plant, the proposed framework isolates the critical parts of the tomato fruit, such as the colour, shape, and size of tomatoes.
The block diagram of the proposed framework is shown in Figure \ref{fig1:block}, where the authors can observe that it is composed of an encoder, transformer, and decoder blocks.

\section{Experiments}
\label{e}
The proposed framework was tested using a dataset from a nearby greenhouse farm in Ajban, Abu Dhabi, UAE.
The dataset comprises time-linked frames that can be employed to identify tomatoes at different maturity levels.
The authors employed meticulous manual annotations using the Matlab data annotations tool to ensure accurate and reliable annotations. Skilled participants used mobile phone cameras to capture tomato images from the greenhouses, and each image was then carefully annotated to identify the ripening stage and other relevant attributes. This annotation process guarantees high-quality and precise labelling, making KUTomaDATA suitable for various computer vision tasks.
Specifically, the authors annotated approximately 700 images of the KUTomaData dataset for three maturity levels, i.e., Unripped, Half ripened and Full ripened tomatoes, and Table \ref{dataset-1} indicates the number of occurrences of each class in this dataset.

\begin{figure*}
     \centering
     \begin{subfigure}[b]{0.7\textwidth}
         \centering
         \includegraphics[width=\textwidth]{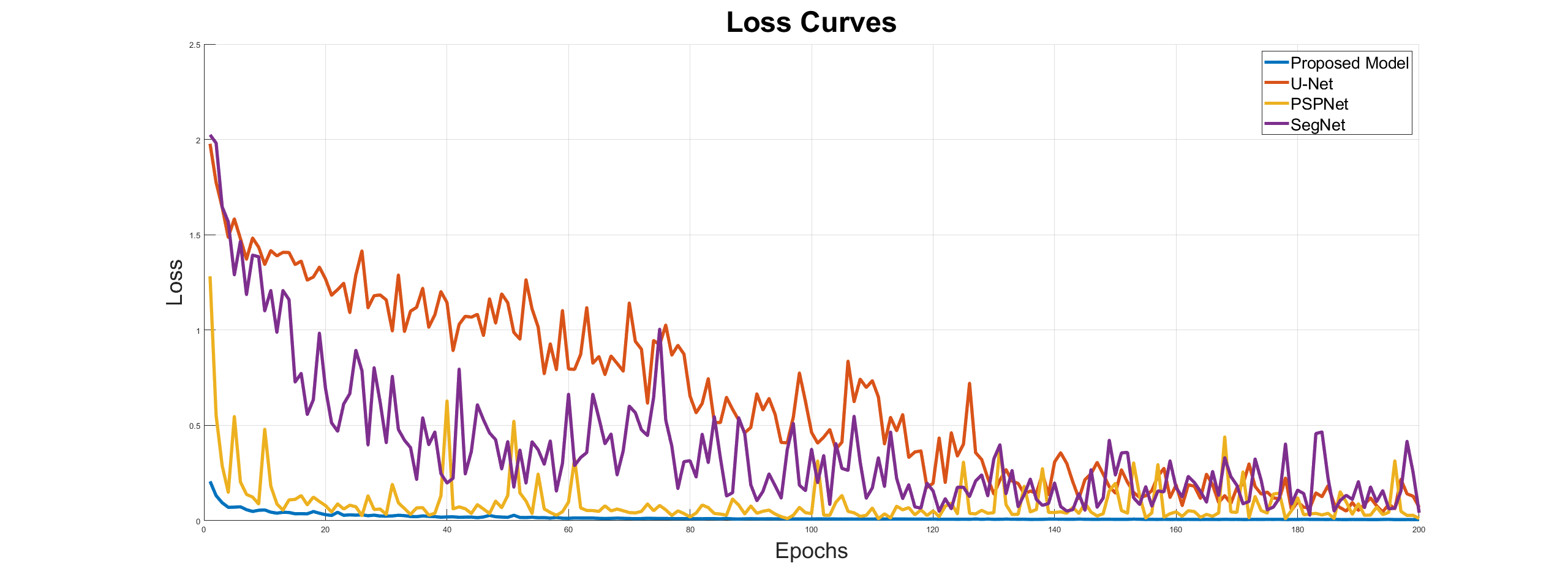}
         \caption{Loss curves for Proposed Model (Our), UNet, PSPNet, and SegNet.}
         \label{fig:Loss}
     \end{subfigure}
     \hfill
     \begin{subfigure}[b]{0.7\textwidth}
         \centering
         \includegraphics[width=\textwidth]{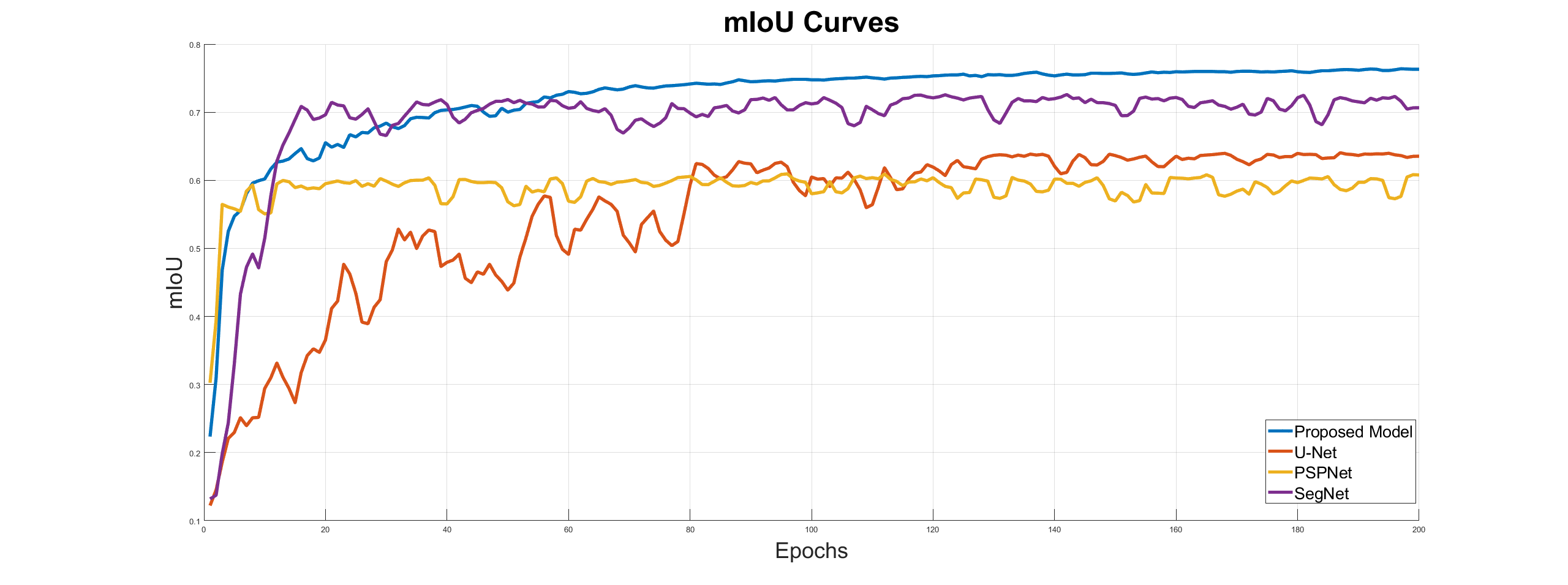}
         \caption{Accuracy curves for Proposed Model (Our), UNet, PSPNet and SegNet. }
         \label{fig:Acc}
     \end{subfigure}
     \hfill
     \caption{Sub-figures (a) and (b) depict the Loss and Accuracy curves, respectively, for several network models during both training and validation stages. The models include the Proposed Model (Our), UNet, PSPNet, and SegNet. }
     \label{loss-miou}
     \end{figure*}

To ensure model robustness, 75\% of the annotated images were used for training, whereas the remaining 25\% was allocated for validation and testing.
During the training phase, the number of epochs and the batch size were set to 200 and 16, respectively.
After each epoch, the trained model was evaluated against the validation dataset. The loss and mIoU curves are presented in Figure~\ref{loss-miou}.
 
     \begin{table}[ht]
\centering
\caption{The proposed model was trained, tested, and evaluated using the KUTomaData dataset, which consists of images for each class: unripened tomatoes, half-ripened tomatoes, and full-ripened tomatoes. The respective number of occurrence in all the images is mentioned here.}
\begin{tabular}{ccc}
\hline
\textbf{Class label} & \textbf{No. of occurrences}\\ \hline
Unripened tomato          & 3557 \\ \hline
Half ripen tomato           & 196 \\  \hline
Full ripen tomato          & 724\\  \hline

\end{tabular}
\label{dataset-1}
\end{table}

Numerous experiments were carried out to assess the proposed method's effectiveness. One of these experiments involved using a test set to assess the model's ability to make accurate predictions under various lighting conditions, occlusion levels, and viewing angles.
The segmentation quality was evaluated by calculating the Dice coefficient and the mean intersection over union (mIoU). These metrics assessed the accuracy and overlap between the predicted segmentation masks and the ground truth annotations.
The Dice coefficient and mean IoU are valuable metrics in gauging the performance and quality of segmentation algorithms, offering complementary insights into the correctness and overlap of the segmentation results. \AK{To evaluate models using mAP, we employ a method where we build bounding boxes from semantic segmentation ground truths and compute minimum bounding rectangles around segmented objects. This technique ensures the creation of bounding boxes precisely fitting the geometry of segmented objects by determining the smallest rectangle containing the entire object. Bounding box limits are determined by finding the least and highest row and column indices where the segmentation mask is 'True,' accurately depicting identified regions in space by closely following segmented object contours. We extend this approach to evaluate both the bounding boxes of the model's output and the mAP from the bounding boxes of the ground truths and the test dataset.}
\MS{
In this work, we compute mAP scores directly from segmentation masks rather than bounding boxes (detections). For each segmentation mask, we extract its minimum and maximum extents to obtain x, y, width, and height information, allowing us to fit a bounding box around the mask. Additionally, we calculate the average confidence scores of each mask pixel and use the mask label, average confidence score, and the fitted bounding box (derived from the segmentation mask) to compute the mAP score. As the mAP score is directly derived from the segmentation mask, the quality of the segmentation mask significantly influences the computed mAP score.
}


\subsection{Experimental Setup:}

The suggested framework has been trained on a system comprising a Core i9-10940 processor running at 3.30 GHz, with 128GB of RAM, and a single NVIDIA Quadro RTX 6000 GPU. The GPU has the CUDA toolkit version 11.0 and cuDNN version 7.5. The development of the proposed model was carried out using Python 3.7.9 and TensorFlow 2.1.0. During the training process, the model was trained for 200 epochs, each consisting of 512 iterations. The ADADELTA optimizer was employed, utilizing default values for the learning rate (1.00) and decay rate (0.95).

\subsection{Data Augmentation:}

\begin{figure*}[h!]
\centering
\includegraphics[width=0.9\textwidth]{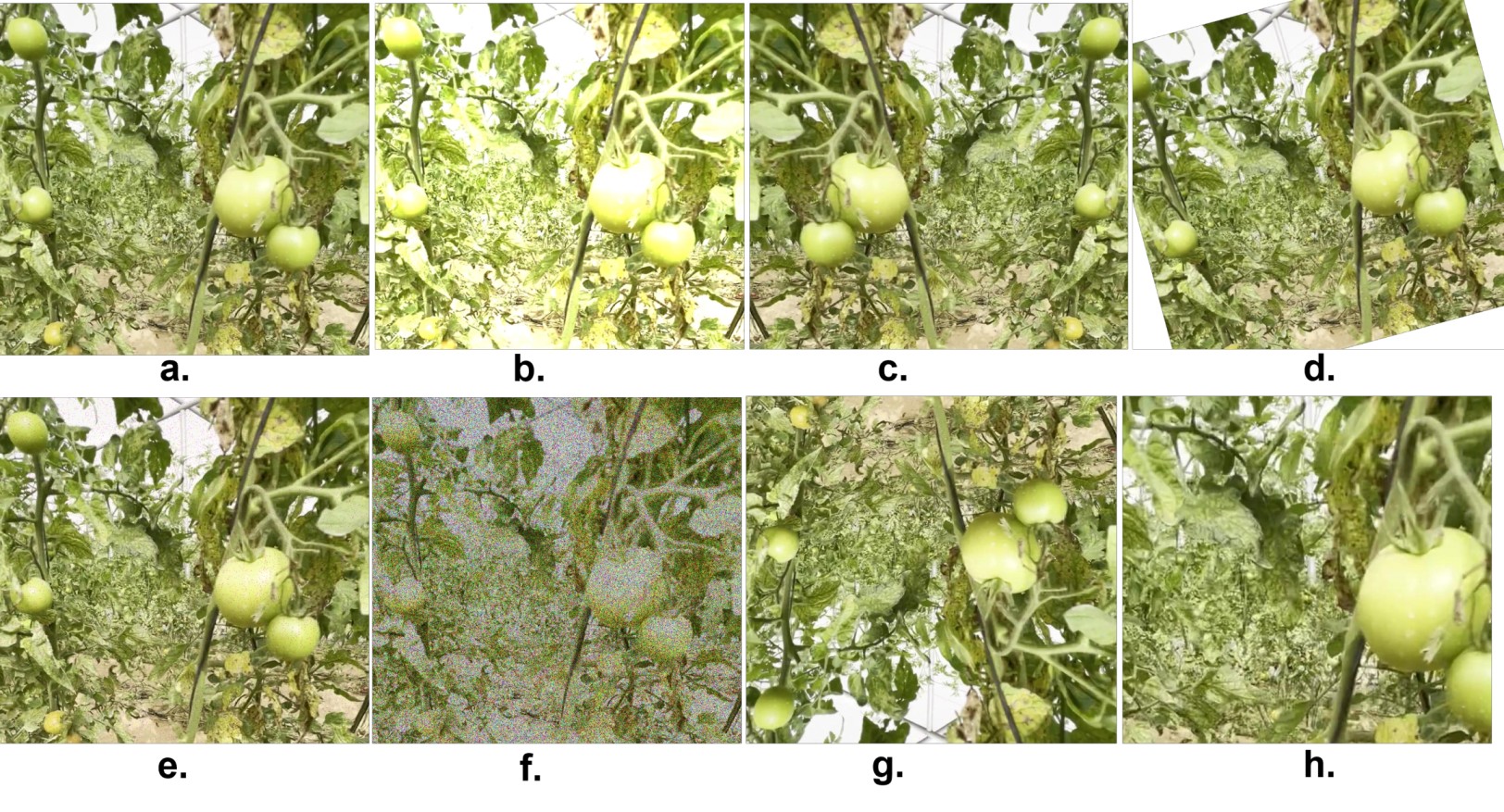}
\caption{Here are some examples of data augmentation techniques: (a). Original image, (b). Random brightness, (c) Horizontal flip, (d). Random rotation, (e). Salt \& pepper, (f). Speckle effect, (f). Vertical variation, and (f). Zoom variation. 
}
\label{fig:aug}
\end{figure*}

Deep Convolutional Neural Network (DCNN) models typically require a substantial number of training images to achieve high accuracy in predicting ground truth labels.
However, there are instances where certain classes may have limited images, posing a challenge in effectively training the model.
Data augmentation techniques are employed to augment the available images and expand the training dataset to tackle this issue.
In our study, the authors employed data augmentation techniques, as described in \cite{shorten2019survey}, to generate additional variations from the existing images for classes with limited samples, particularly for maturity-level classes.
These augmentation techniques include blurriness, rotation, horizontal and vertical flipping, horizontal and vertical shearing, and adding noise.
Figure \ref{fig:aug} illustrates an example of image augmentation. By incorporating this technique, the authors increased the number of images in our dataset, thereby enhancing the model's robustness during the training phase of the CNN.

\section{Results}
\label{r}

In this section, the authors present both qualitative and quantitative results of our experiments, evaluating the performance of each model using several key metrics, including Intersection over Union ($\mu$IoU), Dice coefficient ($\mu$DC), mean Average Precision (mAP), and Area Under the Curve (AUC). These evaluation metrics provide comprehensive insights into the effectiveness and accuracy of the models in various aspects. 

In the following section, an explanation of the theoretical aspects related to network selection is provided. This sub-section aims to provide a comprehensive understanding of how network selection was done underlying the principles involved in the process.

\subsection{Comparison with Conventional Segmentation Models:}

\begin{figure*}[h!]
\centering
\includegraphics[width=0.98\textwidth]{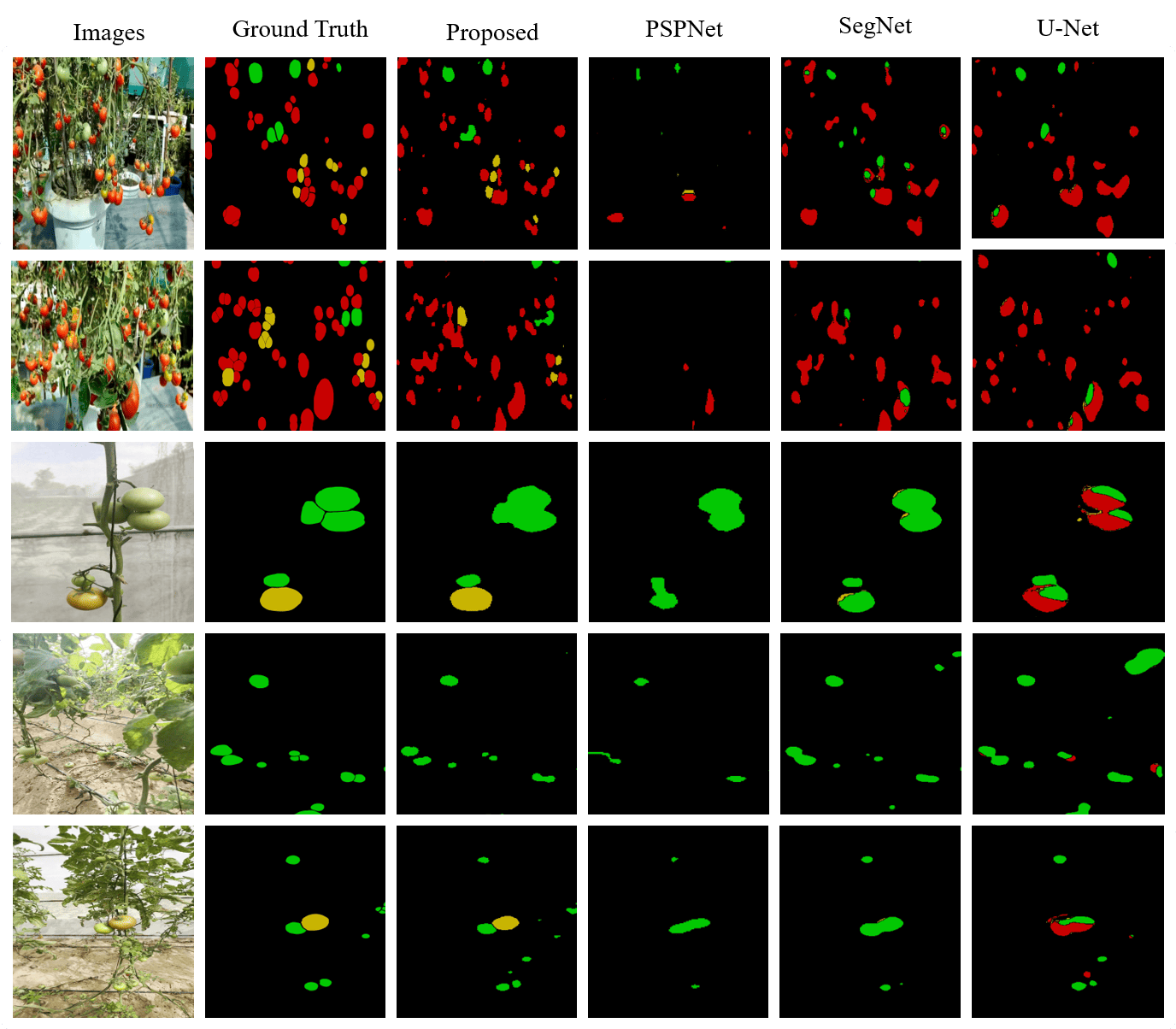}
\caption{The authors compared the proposed framework with the best existing models to evaluate how well it would work. Here, the raw test images from our dataset are displayed in Column 1, the ground truth labels are displayed in Column 2, the results of the proposed framework are displayed in Column 3, and those of PSPNet, SegNet, and UNet are displayed in Columns 4-6, respectively.}
\label{fig:results_cluttered}
\end{figure*}

Figure \ref{fig:results_cluttered} shows tomatoes in a cluttered and occluded environment where the difficulty lies in detecting the unripened tomatoes within same-coloured leaves. This presents a scenario where mobile robots can capture the image and identify the tomatoes. The authors thoroughly assess the proposed framework on the collected dataset. 
Furthermore, the authors also report its comparative evaluation with state-of-the-art segmentation models. Figure \ref{fig:results_cluttered} shows the cluttered situation in an indoor greenhouse where multiple tomato vines can be seen. 
Moreover, the qualitative evaluation of the proposed architecture and its comparison with the state-of-the-art segmentation models (such as SegFormer \cite{xie2021segformer}, PSPNet \cite{zhao2017pyramid}, SegNet \cite{badrinarayanan2017segnet} and U-Net \cite{ronneberger2015unet}) on the dataset is presented in Figure \ref{fig:results_cluttered}.

Table \ref{tab:seg_perfor} represents the quantitative performance of the proposed framework compared to the state-of-the-art networks. It can be seen that the proposed model outperforms the other models in terms of evaluation metrics.
The proposed incremental instance segmentation scheme was compared with various popular transformer, scene parsing, encoder-decoder, and fully convolutional-based models, such as SegFormer \cite{xie2021segformer}, SegNet \cite{badrinarayanan2017segnet}, U-Net \cite{ronneberger2015unet}, and PSPNet \cite{zhao2017pyramid}. As shown in Table \ref{tab:seg_perfor}, this table compares the performance metrics of four different models (Proposed (Our), SegFormer, SegNet, U-Net, and PSPNet) on the task of tomato segmentation. 

The compared metrics are F1 Score, Dice Coefficient, mean Intersection over Union (IoU), and class-wise IoU. The Dice Coefficient is a statistical measure of the overlap between two sets of data - in this case, the predicted and actual tomato segmentation masks. The Mean IoU measures how well the model can accurately segment the tomato regions in the images. The class-wise IoU shows the IoU score for each of the three tomato ripeness classes: Unripe, Half-Ripe, and Fully-Ripe.
The results show that the proposed model outperforms other models in all metrics, achieving a Dice coefficient of 0.7326 and a mean IoU of 0.6641. The proposed model also achieves higher class-wise IoU scores for all three tomato ripeness classes, indicating that it is better at accurately segmenting each class.

The proposed model outperformed SegFormer, SegNet, U-Net, and PSPNet models across all metrics. The SegNet and U-Net models exhibited significantly poorer performance, achieving the lowest scores in all metrics. Their Dice Coefficients were 0.5728 and 0.5475, and mean IoU values were merely 0.4104 and 0.3769, respectively.
The data suggest that the proposed model is highly effective in accurately segmenting tomato regions in images, outperforming other commonly used segmentation models for this particular task.
In addition to quantitative evaluations, a qualitative comparison was performed between the proposed convolutional transformer segmentation framework and other existing segmentation models. The results, illustrated in Figure \ref{fig:results_cluttered}, demonstrate that while all the examined segmentation models successfully localize tomato data through masks, substantial variation exists in the quality of the generated masks across different methods. Notably, our proposed framework exhibits exceptional accuracy in producing precise tomato masks.

Moreover, when considering the extraction of tomatoes at various maturity levels, the capabilities of the proposed convolutional transformer model become evident. Our framework stands out due to its distinctive ability to generate shape-preserving embeddings and to effectively leverage self-attention projections. This unique attribute enables the framework to achieve effective segmentation, even in the presence of occluded tomato data, surpassing the performance of state-of-the-art methods in this domain.

\begin{table}[h!]
\centering
\caption{The performance of the proposed framework was compared to state-of-the-art frameworks using the KUTomaData dataset. The leading results are highlighted in bold, while the second-best scores are underlined.}
\label{tab:seg_perfor}
\begin{tabular}{c|c|c|c|cc}
\hline
\multirow{2}{*}{\textbf{Model}} & \multirow{2}{*}{\textbf{$\mu$DC}} & \multirow{2}{*}{\textbf{$\mu$IoU}} & \multicolumn{3}{c}{\textbf{Classwise IoU}} \\ \cline{4-6} 
                                &                                           &                                    & \multicolumn{1}{c|}{\textit{Unriped}} & \multicolumn{1}{c|}{\textit{Half-Riped}} & \textit{Fully Riped} \\ \hline
Proposed (\textbf{Our})         & \textbf{0.7685}                           & \textbf{0.6241}                    & \multicolumn{1}{c|}{\textbf{0.7395}}  & \multicolumn{1}{c|}{\textbf{0.6028}}    & \textbf{0.3262}      \\ \hline
SegFormer \cite{xie2021segformer}                      & { 0.7297}                           & { 0.5745}                    & \multicolumn{1}{c|}{{ 0.6391}}  & \multicolumn{1}{c|}{ 0.3969}    & { 0.2800}    \\ \hline
SegNet \cite{badrinarayanan2017segnet}                          & {0.5728}                              & { 0.4104}                       & \multicolumn{1}{c|}{{ 0.6288}}     & \multicolumn{1}{c|}{0.0001}             & 0.0002               \\ \hline
UNet \cite{ronneberger2015unet}                           & 0.5475                                    & 0.3769                             & \multicolumn{1}{c|}{0.5422}           & \multicolumn{1}{c|}{0.0016}             & {0.0017}         \\ \hline
PSPNet \cite{zhao2017pyramid}                          & 0.5504                                    & 0.3797                             & \multicolumn{1}{c|}{0.5123}           & \multicolumn{1}{c|}{{0.0320}}       & 0.0001               \\ \hline
\end{tabular}
\end{table}

\subsection{Quantitative Evaluations:}

Table \ref{tab:seg_perfor} presents a quantitative comparison of different models based on various evaluation metrics for tomato segmentation. These metrics include the Dice Coefficient, Mean IoU (Intersection over Union), and Classwise IoU (IoU for different tomato ripeness classes). The first row represents the proposed model, labelled as "Our", which achieved a remarkably high Dice Coefficient of 0.7326 and a mean IoU of 0.6641. The Classwise IoU values for the "Unripened," "Half-Ripened," and "Fully Ripened" classes are also noteworthy, with IoU scores of 0.7395, 0.6028, and 0.3262, respectively. Comparing the proposed model to other state-of-the-art models, a Dice Coefficient of 0.6602 and a mean IoU of 0.5745. However, its Classwise IoU scores for all three ripeness classes are lower than the proposed model's. The SegNet model obtained a Dice Coefficient of 0.5728 and a mean IoU of 0.4104. Its Classwise IoU scores for the "Unripened" and "Half-Ripened" classes are higher than those of other models, but it performs poorly for the "Fully Ripened" class. The UNet model achieved a Dice Coefficient of 0.5475 and a mean IoU of 0.3769. Similar to SegNet, it demonstrates better performance for the "Unripened" and "Half-Ripened" classes but struggles with the "Fully Ripened" class.

Finally, the PSPNet model obtained a Dice Coefficient of 0.5504 and a Mean IoU of 0.3797. Its Classwise IoU scores for the "Unripened" and "Half-Ripened" classes are relatively higher, but it performs poorly for the "Fully Ripened" class. Overall, the proposed model outperforms the other models regarding the Dice Coefficient, mean IoU, and Classwise IoU for different tomato ripeness classes. The results highlight the effectiveness and superiority of the proposed model in accurately segmenting tomatoes of varying ripeness levels.

\begin{table}[t]
\footnotesize
    \centering
    \caption{ Quantitative evaluation of the proposed framework with state-of-the-art methods in terms of $\mu$IoU, $\mu$DC, mAP, and AUC scores across KUTomaData, Laboro, and Rob2Pheno datasets.}
    \begin{tabular}{cccccccc}
         \hline
         Dataset & Method & $\mu$IoU& $\mu$DC & mAP & AUC \\\hline
         KUTomaData & Proposed & 0.6241 & 0.7685 & 0.5814 & 0.7381 \\
         & SETR \cite{setr} & 0.5923 & 0.7439 & 0.5382 & 0.7069 \\
         & SegFormer \cite{xie2021segformer} & 0.5745 & 0.7297 & 0.5176 & 0.6843 \\
         & DeepFruits \cite{deepfruits} & 0.4668 & 0.6364 & 0.4368 & 0.6209 \\
         & COS \cite{COS} & 0.4837 & 0.6520 & 0.4562 & 0.5968 \\
         & CWD \cite{CWD} & 0.5096 & 0.6751 & 0.4739 & 0.6027 \\
         & DLIS \cite{Horticulture} & 0.5485 & 0.7084 & 0.5173 & 0.6391 \\\hline

         Laboro & Proposed & 0.6946 & 0.8197 & 0.6542 & 0.7419 \\
         & SETR \cite{setr} & 0.6529 & 0.7900 & 0.6083 & 0.7346 \\
         & SegFormer \cite{xie2021segformer} & 0.6387 & 0.7795 & 0.5856 & 0.7068 \\
         & DeepFruits \cite{deepfruits} & 0.5162 & 0.6809 & 0.4602 & 0.5834 \\
         & COS \cite{COS} & 0.5243 & 0.6879 & 0.4728 & 0.5812 \\
         & CWD \cite{CWD} & 0.5576 & 0.7159 & 0.5116 & 0.6185 \\
         & DLIS \cite{Horticulture} & 0.5865 & 0.7393 & 0.5394 & 0.6527 \\\hline
         
         Rob2Pheno & Proposed & 0.7341 & 0.8466 & 0.6639 & 0.8253 \\
         & SETR \cite{setr} & 0.6856 & 0.8134 & 0.6204 & 0.7261 \\
         & SegFormer \cite{xie2021segformer} & 0.6738 & 0.8151 & 0.6325 & 0.7524 \\
         & DeepFruits \cite{deepfruits} & 0.5967 & 0.7474 & 0.5315 & 0.6403 \\
         & COS \cite{COS} & 0.6149 & 0.7615 & 0.5628 & 0.6752 \\
         & CWD \cite{CWD} & 0.6424 & 0.7822 & 0.5935 & 0.7124 \\
         & DLIS \cite{Horticulture} & 0.6573 & 0.7932 & 0.6176 & 0.7492 \\\hline
         
    \end{tabular}
    \label{tab:quan}
\end{table}

\subsection{Qualitative Evaluations:}

\begin{figure*}[t]
\centering
\includegraphics[width=1\textwidth]{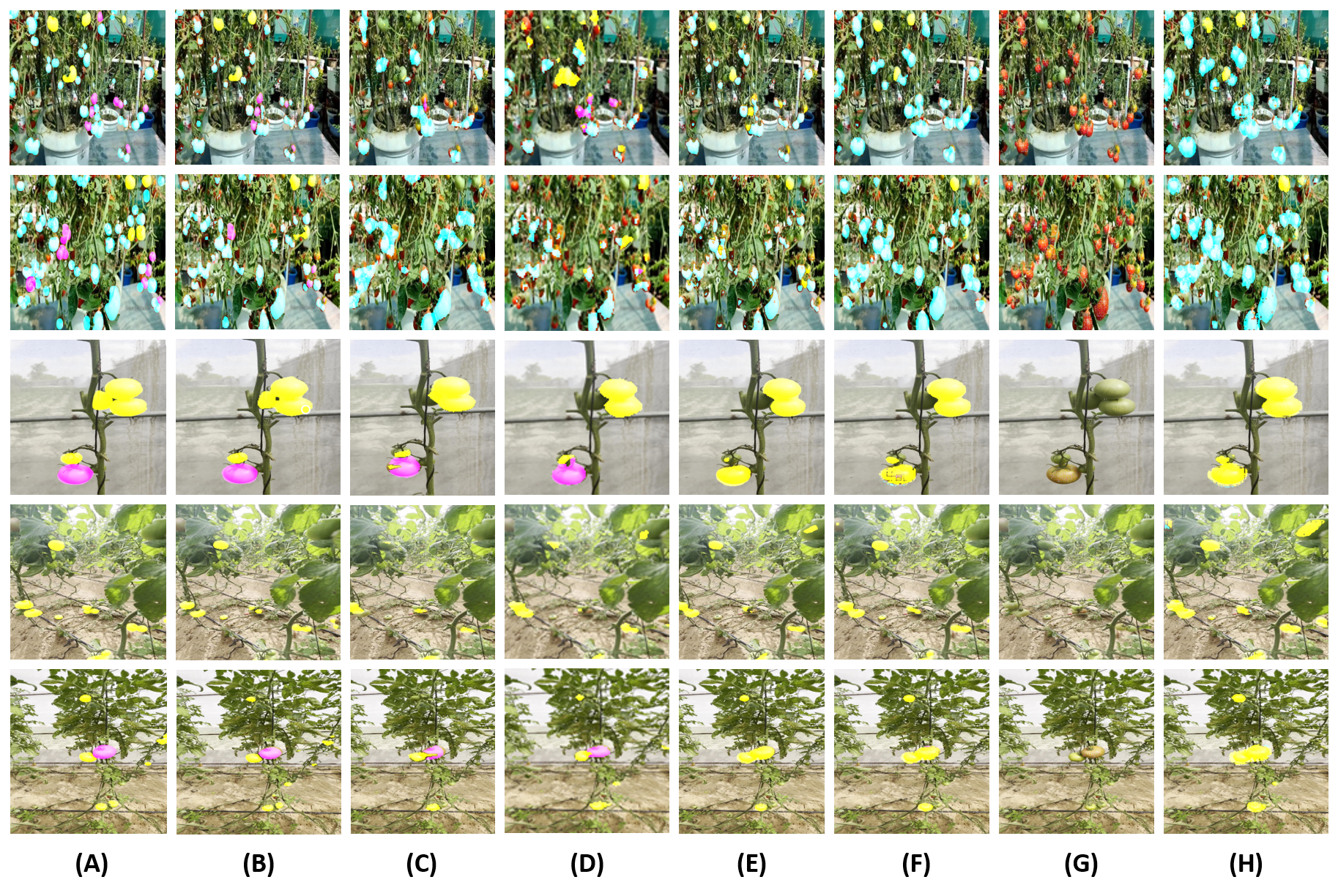}
\caption{Qualitative evaluation of the proposed framework alongside state-of-the-art methods to extract different maturity grades of tomatoes under occlusion and variable lighting conditions. Column (A) represents the ground truth overlaid on the actual image, where cyan represents ripe tomatoes, pink represents half-ripe tomatoes, and yellow highlights unripe tomatoes. Column (B) shows the outcome of the proposed method, while Columns (C), (D), (E), (F), (G), and (H) display the qualitative analysis for SETR \cite{setr}, Segformer \cite{xie2021segformer}, DeepFruits \cite{deepfruits}, COS \cite{COS}, CWD \cite{CWD} and DLIS \cite{Horticulture}, respectively.
}
\label{fig:qual}
\end{figure*}

Figure~\ref{fig:qual} presents a rigorous qualitative assessment of the proposed framework alongside state-of-the-art methods, primarily focusing on the accuracy of tomato segmentation. The objective is to comprehensively evaluate the performance of the proposed framework against existing approaches when dealing with real-world scenarios.The quantitative analysis of these models is shown in Table \ref{tab:quan}.

In Column (A) of Figure \ref{fig:qual}, the ground truth annotations are visually overlaid on the corresponding actual images. A distinctive colour scheme is employed to signify different maturity grades: cyan for fully-ripened tomatoes, pink for half-ripened tomatoes, and yellow for unripe tomatoes. This column is a reliable reference for assessing the expected quality of segmentation. Column (B) showcases the exceptional results of the proposed convolutional transformer model. The segmentation outcomes achieved by the framework demonstrate its remarkable efficacy in accurately classifying and segmenting tomatoes of three maturity grades, even in scenarios with challenging factors such as occlusion and variable lighting conditions. Columns (C) to (H) provide a meticulous comparative analysis of other state-of-the-art methods, namely SETR \cite{setr}, Segformer \cite{xie2021segformer}, DeepFruits \cite{deepfruits}, COS \cite{COS}, CWD \cite{CWD}, and DLIS \cite{Horticulture}. Each column represents a distinct method, illustrating the segmentation results attained by the respective approaches. 
This thorough evaluation facilitates a meticulous examination and meaningful comparisons of the techniques, leading to the identification of the most effective segmentation model for tomatoes.

It is also evident from Figure \ref{fig:qual} that the proposed framework consistently outperforms state-of-the-art methods in accurately extracting tomatoes of different maturity grades. The segmentation results obtained by the proposed method exhibit superior accuracy, robustness, and the ability to precisely classify and delineate fully-riped, half-riped, and unripe tomatoes, even in challenging conditions. Conversely, the qualitative analysis of alternative methods reveals varying performance levels, with specific approaches struggling to delineate the distinct maturity grades accurately.

\section{Ablation Study}
 \label{as}

The authors conduct an ablation study in this section to pinpoint the optimal hyperparameters and backbone networks that yield the most favourable outcomes across various datasets.
The first set of ablation experiments focused on identifying optimal $\beta$ parameters that produce the best recognition performance of the proposed framework.
The second set of experiments aimed to identify the optimal network backbone.
Several backbone architectures were evaluated and compared to discern the architecture that yielded maximum accuracy and quality segmentation. 
The objective of the third series of experiments was to determine the optimal value for the parameter $\tau$.
By varying $\tau$ and evaluating the model's performance, the authors established the threshold that maximized detection accuracy while minimizing false positives and negatives.
The fourth ablation experiment aimed to identify the optimal loss function for the proposed model by comparing it to other state-of-the-art loss functions, including soft nearest neighbour loss, focal Tversky loss, dice-entropy loss, and conventional cross-entropy loss.  
The fifth series of ablation experiments was related to comparing the segmentation performance of the proposed model against state-of-the-art networks.

\subsection{Optimal  \texorpdfstring{$\beta$ values in $L_{t}$}{}}

The first set of ablation experiments aimed to determine the optimal hyper-parameters $\beta_{1,2}$ in the $L_t$ loss function, which would result in the best segmentation performance across different datasets. \AK{To explore this, the authors varied the value of $\beta_1$ from 0.1 to 0.9 in increments of 0.2. For each $\beta_1$ value, the authors calculated $\beta_2$ as $\beta_2=1-\beta_1$.} Subsequently, the proposed model was trained using each combination of $\beta_1$ and $\beta_2$. During the inference stage, the model's segmentation performance for each combination was evaluated across the datasets, utilizing mAP scores as the evaluation metric (as shown in Table \ref{tab:beta}).

The results revealed that the proposed framework performs better when assigning a higher weight to $\beta_1$, particularly with a value of 0.9 in this specific instance. For example, with $\beta_1=0.9$ and $\beta_2=0.1$, the proposed model achieved mAP scores of 0.5814, 0.6542, and 0.6639 across the three datasets: KUTomaData, Laboro Tomato, and Rob2Pheno Annotated Tomato respectively. Based on these findings, a combination of $\beta_1=0.9$ and $\beta_2=0.2$ was selected for subsequent experiments to train the proposed model. This choice of hyperparameters was deemed optimal based on the earlier evaluations and resulted in favourable model performance.
\begin{table}[t]
\footnotesize
    \centering
    \caption{ The objective is to determine the optimal values of $\beta_{1,2}$ that yield the highest segmentation performance. In addition, the abbreviations used in the analysis are as follows: DS (Dataset) and KUTomaData(KUTD). The bold values indicate the best performance in terms of mAP (mean Average Precision) score, while '-' signifies that the corresponding combination of $\beta_i$ and $\beta_j$ is not valid or applicable.}
    \begin{tabular}{c | c | c c c c c c}
         \hline
        
         KUTomaData & $\beta_1$ & 0.1 & 0.3 & 0.5 & 0.7 & 0.9  \\\hline
         $\beta_2$ & 0.1 & - & - & - & - & \textbf{0.5814} \\
         & 0.3 & - & - & - & 0.5352 & - \\
         & 0.5 & - & - & 0.4836 & - & - \\
         & 0.7 & - & 0.4582 &- & - & - \\
         & 0.9 & 0.4031 & - & - & - & - \\\hline

         Laboro & $\beta_1$ & 0.1 & 0.3 & 0.5 & 0.7 & 0.9  \\\hline
         $\beta_2$ & 0.1 & - & - & - & - & \textbf{0.6542} \\
         & 0.3 & - & - & - & 0.5894 & - \\
         & 0.5 & - & - & 0.5351 & - & - \\
         & 0.7 & - & 0.4953 &- & - & - \\
         & 0.9 & 0.4406 & - & - & - & - \\\hline

         Rob2Pheno & $\beta_1$ & 0.1 & 0.3 & 0.5 & 0.7 & 0.9  \\\hline
         $\beta_2$ & 0.1 & - & - & - & - & \textbf{0.6639} \\
         & 0.3 & - & - & - & 0.6074 & - \\
         & 0.5 & - & - & 0.5692 & - & - \\
         & 0.7 & - & 0.5106 &- & - & - \\
         & 0.9 & 0.5739 & - & - & - & - \\\hline
         
    \end{tabular}
    \label{tab:beta}
\end{table}

\subsection{Optimal Encoder Backbone}
\begin{sloppypar}
 
The second set of ablation experiments aimed to determine the optimal network backbone for segmenting and detecting tomato objects. \AK{The model has been designed to effectively integrate with several convolutional neural network (CNN) backbones for the encoder.}
To achieve this, the authors integrated various pre-trained models, including HRNet \cite{wang2020deep}, Lite-HRNet \cite{yu2021litehrnet}, EfficientNet-B4 \cite{Tan2019EfficientNetRM}, DenseNet-201 \cite{huang2018densely}, and ResNet-101 \cite{he2015deep}, into the proposed model. 
The authors then compared their performance against the proposed backbone specifically designed for tomato object detection and segmentation. 
The results obtained from the conducted experiments are displayed in Table \ref{tab:backbone}. 
Upon examining Table \ref{tab:backbone}, the proposed encoder outperformed the state-of-the-art models, surpassing them by 3.22\%, 2.51\%, 3.67\%, and 0.56\% in terms of $\mu$IoU, $\mu$DC, mAP, and AUC scores, respectively, on the KUTomaData dataset.

Moreover, when considering the Laboro dataset, the proposed framework exhibited performance improvements of 2.27\%, 1.60\%, 2.61\%, and 2.25\% in terms of $\mu$IoU, $\mu$DC, mAP, and AUC scores, respectively. 
Similarly, on the Rob2Pheno dataset, the proposed model achieved gains of 3.68\%, 2.50\%, 3.71\%, and 1.25\% in $\mu$IoU, $\mu$DC, mAP, and AUC scores, respectively. 
These notable performance improvements can be attributed to utilizing a novel butterfly structure in the proposed encoder backbone. 
\AK{In contrast to traditional encoders, the proposed encoder can maintain the high-resolution features of the candidate input by summing feature maps across each depth in a butterfly manner via upsampling and downsampling the kernel sizes as needed. Moreover, each block within the proposed network consists of custom identity blocks (IB), hierarchical decomposition blocks (HDB), and shape-preservation blocks (SPB). These blocks refine the attention of the model so that it only focuses on the defected regions, irrespective of the scan’s textural and contextual attributes.}
The model acquires the ability to extract distinctive latent characteristics from the input images by adding this integration, resulting in improved performance in tomato object segmentation and classification tasks. This advancement outperforms the capabilities of current cutting-edge models, such as HRNet \cite{wang2020deep}, Lite-HRNet \cite{yu2021litehrnet}, EfficientNet-B4 \cite{Tan2019EfficientNetRM}, DenseNet-201 \cite{huang2018densely}, and ResNet-101 \cite{he2015deep}. It is important to note that while the proposed scheme is computationally expensive compared to Lite-HRNet \cite{yu2021litehrnet}, its superior detection performance justified its selection for generating distinct feature representations in the subsequent experiments. 
This decision was driven by the primary objective of achieving the highest possible detection performance.

\begin{table}[t]
\footnotesize
    \centering
    \caption{To identify the most suitable backbone network for performing tomato object detection and segmentation tasks across all datasets, a comprehensive evaluation was conducted. The models were evaluated using an input size of 540 $\times$ 640 $\times$ 3. 
    } 
    \begin{tabular}{ccccccc}
         \hline
         Dataset & Backbone & $\mu$IoU& $\mu$DC & mAP & AUC & Params \\\hline
         KUTomaData & Proposed & 0.6241 & 0.7685 & 0.5814 & 0.7381 & 52.4M \\
         & Lite-HRNet \cite{yu2021litehrnet} & 0.5753 & 0.7304 & 0.5365 & 0.7168 & 7.43M \\
         & HRNetv2 \cite{wang2020deep} & 0.5916 & 0.7434 & 0.5447 & 0.7325 & 52.1M \\
         & EfficientNetB4 \cite{Tan2019EfficientNetRM} & 0.5324 & 0.6948 & 0.5102 & 0.6762 & 22.3M \\
         & DenseNet-201 \cite{huang2018densely} & 0.5672 & 0.7238 & 0.5283 & 0.6831 & 85.6M \\
         & ResNet-101 \cite{he2015deep} & 0.5384 & 0.6999 & 0.4951 & 0.6676 & 84.2M \\ \hline

         Laboro & Proposed & 0.6946 & 0.8197 & 0.6542 & 0.7498 & 54.3M \\
         & Lite-HRNet \cite{yu2021litehrnet} & 0.6653 & 0.7990 & 0.6173 & 0.7208 & 8.94M \\
         & HRNetv2 \cite{wang2020deep} & 0.6719 & 0.8037 & 0.6281 & 0.7483 & 51.2M \\
         & EfficientNetB4 \cite{Tan2019EfficientNetRM} & 0.6582 & 0.7938 & 0.6017 & 0.7219 & 23.6M \\
         & DenseNet-201 \cite{huang2018densely} & 0.6625 & 0.7969 & 0.6246 & 0.7394 & 87.5M \\
         & ResNet-101 \cite{he2015deep} & 0.6503 & 0.7880 & 0.6128 & 0.7153 & 86.5M \\ \hline
         
        Rob2Pheno & Proposed & 0.7341 & 0.8466 & 0.6639 & 0.8253 & 58.3M \\
         & Lite-HRNet \cite{yu2021litehrnet} & 0.6824 & 0.8112 & 0.6143 & 0.8029 & 9.72M \\
         & HRNetv2 \cite{wang2020deep} & 0.6973 & 0.8216 & 0.6268 & 0.8128 & 53.7M \\
         & EfficientNetB4 \cite{Tan2019EfficientNetRM} & 0.6782 & 0.8082 & 0.6042 & 0.7953 & 22.1M \\
         & DenseNet-201 \cite{huang2018densely} & 0.6856 & 0.8134 & 0.6194 & 0.8058 & 86.4M \\
         & ResNet-101 \cite{he2015deep} & 0.6675 & 0.8006 & 0.5918 & 0.7942 & 85.3M \\ \hline
         
         \hline
    \end{tabular}
    \label{tab:backbone}
\end{table}

\end{sloppypar}

\subsection{Determining the Optimal Temperature Constant}  

In the proposed $L_t$ loss function, the temperature constant ($\tau$) serves as a hyperparameter that softens the target probabilities. 
Using a higher value of $\tau$, the model becomes more receptive to recognising tomato object segmentation and detection regardless of the input imagery characteristics. 
This softening effect enhances the detection and segmentation performance by enabling the model to comprehend the target probabilities more broadly.

In the fourth set of ablation experiments, the authors aimed to determine the optimal value for $\tau$ to extract tomato objects accurately. 
To achieve this, the authors varied the value of $\tau$ from 1 to 2.5 in increments of 0.5 within the $L_t$ loss function while training the proposed model across each dataset. 
After completing the training process, in the inference stage, the authors assessed the performance of the proposed framework in tomato object segmentation and detection on each dataset. 
The outcomes of these evaluations are showcased in the provided table \ref{tab:tau}.
From Table \ref{tab:tau}, it can be observed that increasing the value of $\tau$ from 1 to 1.5 led to a significant performance boost across all four datasets. 
For instance, on the KUTomaData dataset, the proposed framework achieved performance improvements of 4.12\% in terms of $\mu$IoU, 3.21\% in terms of $\mu$DC, 1.65\% in terms of mAP, and 1.25\% in terms of AUC scores. 
Similarly, on the Laboro dataset, it achieved performance improvements of 2.87\% in $\mu$IoU, 2.03\% in $\mu$DC, 3.58\% in mAP, and 1.88\% in AUC scores. Furthermore, experiments on the Rob2Pheno Annotated dataset showed performance improvements of 1.88\% in $\mu$IoU, 1.26\% in $\mu$DC, 2.12\% in mAP, and 1.85\% in AUC scores.

It is important to note that increasing $\tau$ does not always result in performance improvements. When the authors increased the value of $\tau$ from 1.5 to 2 and from 2 to 2.5, the proposed framework's effectiveness deteriorated.
This decline in performance can be attributed to the fact that when $\tau$ exceeds a certain threshold, it loses its ability to accurately differentiate between logits representing different categories, such as green, half-ripen and fully-ripen and the background, within the input imagery.

Considering the optimal detection results achieved with $\tau=1.5$ for the proposed framework on each dataset, the authors chose to train the model with $\tau=1.5$ for the remaining experiments. 
This selection ensures consistent and effective performance throughout the subsequent experimentation.

\begin{table}[t]
\footnotesize
    \centering
    \caption{To identify the optimal temperature constant ($\tau$) for achieving the best tomato detection performance across each dataset, the authors employed the proposed backbone encoder to generate latent features using various values of $\tau$.
   } 
  
    \begin{tabular}{cccccc}
         \hline
         Dataset & $\tau$ & $\mu$IoU& $\mu$DC & mAP & AUC \\\hline
         KUTomaData & 1 & 0.5829 & 0.7364 & 0.5649 & 0.7256 \\
         & 1.5 & 0.6241 & 0.7685 & 0.5814 & 0.7381 \\
         & 2 & 0.5783 & 0.7328 & 0.5627 & 0.7169 \\
         & 2.5 & 0.5596 & 0.7176 & 0.5345 & 0.6812 \\ \hline

         Laboro & 1 & 0.6528 & 0.7899 & 0.6013 & 0.7293 \\
         & 1.5 & 0.6946 & 0.8197 & 0.6542 & 0.7419 \\
         & 2 & 0.6659 & 0.7994 & 0.6184 & 0.7156 \\
         & 2.5 & 0.6407 & 0.7810 & 0.6025 & 0.6924 \\ \hline
         
         Rob2Pheno & 1 & 0.7026 & 0.8253 & 0.6284 & 0.8068 \\
         & 1.5 & 0.7341 & 0.8466 & 0.6639 & 0.8253 \\
         & 2 & 0.7153 & 0.8340 & 0.6427 & 0.7976 \\
         & 2.5 & 0.6938 & 0.8192 & 0.6265 & 0.7782 \\ \hline
         
         \hline
    \end{tabular}
    \label{tab:tau}
\end{table}

\subsection{Optimal Loss Function}

The fifth set of ablation experiments focused on analysing the performance of the proposed model when trained using the $L_t$ loss function compared to other state-of-the-art loss functions. 
These include the soft nearest neighbor loss function ($L_{sn}$) \cite{lsnAnalysis}, the focal Tversky loss function ($L_{ft}$) \cite{lft}, the dice-entropy loss function ($L_{de}$) \cite{lde}, and the conventional cross-entropy loss function ($L_{ce}$). The results of these experiments are summarised in Table \ref{tab:lossFunction}. From Table \ref{tab:lossFunction}, it is evident that the proposed model, trained using the $L_t$ loss function, outperformed its counterparts trained with state-of-the-art loss functions across all datasets. 
For instance, on the KUTomaData dataset, the $L_t$ loss function resulted in a performance improvement of 2.16\% in terms of $\mu$IoU, 1.66\% in terms of $\mu$DC, 2.39\% in terms of mAP, and 2.25\% in terms of AUC scores. 
Similarly, on the Laboro dataset, the $L_t$ loss function led to a performance improvement of 3.25\% in terms of $\mu$IoU, 2.30\% in terms of $\mu$DC, 5.58\% in terms of mAP, and 4.81\% in terms of AUC scores. 

Furthermore, on the Rob2Pheno Annotated dataset, it yielded a performance improvement of 1.23\% in terms of $\mu$IoU, 0.82\% in terms of $\mu$DC, 1.16\% in terms of mAP, and 1.45\% in terms of AUC scores.
 
These performance improvements can be attributed to the proposed $L_t$ loss function, which leverages both contextual and semantic differences within the underwater scans, effectively allowing the model to recognise tomato objects regardless of input image characteristics. 
Consequently, the authors employed the $L_t$ loss function for the remaining experiments to train the proposed model for tomato object extraction across all four datasets.

\begin{table}[t]
\footnotesize
    \centering
    \caption{To determine the optimal loss function with the best tomato object detection performance across all four datasets, the authors used the proposed backbone encoder to generate the latent features when the model was constrained using different loss functions. 
    Moreover, (P) indicates the proposed $L_t$ loss function.
    }
    
    \begin{tabular}{cccccccc}
         \hline
         Dataset & Loss Function & $\mu$IoU& $\mu$DC & mAP & AUC \\\hline
         KUTomaData & $L_t$ (P) & 0.6241 & 0.7685 & 0.5814 & 0.7381 \\
         & $L_{ce}$ & 0.5516 & 0.7110 & 0.5143 & 0.6836 \\
         & $L_{sn}$ \cite{lsnAnalysis} & 0.6025 & 0.7519 & 0.5575 & 0.7156 \\
         & $L_{ft}$ \cite{lft} & 0.5773 & 0.7320 & 0.5164 & 0.7052 \\
         & $L_{de}$ \cite{lde} & 0.5962 & 0.7470 & 0.5389 & 0.7109 \\ \hline

         Laboro & $L_t$ (P) & 0.6946 & 0.8197 & 0.6542 & 0.7419 \\
         & $L_{ce}$ & 0.6183 & 0.7641 & 0.5423 & 0.6537 \\
         & $L_{sn}$ \cite{lsnAnalysis} & 0.6621 & 0.7967 & 0.5984 & 0.6938 \\
         & $L_{ft}$ \cite{lft} & 0.6358 & 0.7773 & 0.5667 & 0.6726 \\
         & $L_{de}$ \cite{lde} & 0.6496 & 0.7875 & 0.5793 & 0.6893 \\ \hline
         
         Rob2Pheno & $L_t$ (P) & 0.7341 & 0.8466 & 0.6639 & 0.8253 \\
         & $L_{ce}$ & 0.6793 & 0.8090 & 0.6128 & 0.7641 \\
         & $L_{sn}$ \cite{lsnAnalysis} & 0.6946 & 0.8197 & 0.6315 & 0.7869 \\
         & $L_{ft}$ \cite{lft} & 0.7104 & 0.8306 & 0.6447 & 0.8076 \\
         & $L_{de}$ \cite{lde} & 0.7218 & 0.8384 & 0.6523 & 0.8108 \\ \hline
         
    \end{tabular}
    \label{tab:lossFunction}
\end{table}

\subsection{Transformer Encoder Analysis}

\AK{In this subsection, we conduct a comprehensive ablation study focused on removing the transformer encoder component from our proposed network architecture. We aim to investigate in detail the impact of the transformer encoder in our fully convolutional pipeline. By systematically evaluating the model's performance with and without the transformer encoder, we aim to clarify its crucial role in enhancing the feature extraction capabilities of the proposed model for our specific task. 

\begin{table}[h!]
\small
    \centering \AK{
    \caption{Comparison of Model Variants with and Without Transformer Encoder Across Datasets} 
  
    \begin{tabular}{cccc}
         \hline
         Dataset & Variant & $\mu$IoU& $\mu$DC  \\\hline
         KUTomaData & Proposed (With Transformer Block) & 0.6241 & 0.7685 \\
         & Proposed (Without Transformer Block) & 0.5938 & 0.7486 \\ \hline

        Laboro & Proposed (With Transformer Block) & 0.6946 & 0.8197 \\
         & Proposed (Without Transformer Block) & 0.6572 & 0.7931 \\ \hline

        Rob2Pheno & Proposed (With Transformer Block) & 0.7341 & 0.8466 \\
         & Proposed (Without Transformer Block) & 0.7059 & 0.8275 \\ \hline
         \hline
    \end{tabular}
    \label{tab:trans_abla}}
\end{table}

From Table~\ref{tab:trans_abla}, it is evident that the model using the transformer encoder showed notable performance gains across KUTomaData, Laboro, and Rob2Pheno. More specifically, the mean Dice Coefficient (mDC) and mean Intersection over Union (mIoU) scores increased with the addition of the transformer block. For instance, using the transformer, the mDC increased by 2.86\% on KUTomaData, and the mIoU improved by 4.13\%. Comparable patterns were noted in the Rob2Pheno and Laboro datasets. These findings support our theory that adding transformer-based designs to fully convolutional pipelines improves the model's capacity to identify complex linkages and patterns in the input. Improved semantic segmentation across datasets results from the transformer's attention mechanisms, which are essential for precise object segmentation.
}

\section{Discussion}
\label{d}

The proposed framework presents a novel approach for tomato maturity level segmentation and classification using RGB scans acquired under various lighting and occlusion conditions. The experimental analysis demonstrates the framework's effectiveness in segmenting and grading tomatoes based on colour, shape, and size.
The proposed framework addresses the challenges associated with harvesting ripe tomatoes using mobile robots in real-world scenarios. These challenges include occlusion caused by leaves and branches and the colour similarity between tomatoes and the surrounding foliage during fruit development. The existing literature lacks a sufficient explanation of these tomato recognition challenges, necessitating the development of new approaches.
To overcome these challenges, a novel framework is introduced in this paper, leveraging a convolutional transformer architecture for autonomous tomato recognition and grading. The framework is designed to handle tomatoes with varying occlusion levels, lighting conditions, and ripeness stages. It offers a promising solution for efficient tomato harvesting in complex and diverse natural environments.
An essential contribution of this work is the introduction of the KUTomaData dataset, specifically curated for training deep learning models for tomato segmentation and classification. KUTomaData comprises images collected from greenhouses across the UAE. The dataset encompasses diverse lighting conditions, viewing perspectives, and camera sensors, making it unique compared to existing datasets. The availability of KUTomaData fills a gap in the deep learning community by providing a dedicated resource for tomato-related research.
The proposed framework's performance was evaluated against two additional public datasets: Laboro Tomato and Rob2Pheno Annotated Tomato. These datasets were used to benchmark the framework's ability to extract cluttered and occluded tomato instances from RGB scans, comparing its performance against state-of-the-art models. The evaluation results demonstrated exceptional performance, with the proposed framework outperforming the state-of-the-art models, including SETR \cite{setr}, Segformer \cite{xie2021segformer}, DeepFruits \cite{deepfruits}, COS \cite{COS}, CWD \cite{CWD}, and DLIS \cite{Horticulture}, by a significant margin.
A series of ablation experiments were conducted to enhance the model's effectiveness. The initial experiments focused on optimizing hyperparameters to improve performance. Subsequently, different network backbones were compared in the second set of experiments to identify the architecture that achieved accurate and high-quality segmentation. The fourth set of experiments determined the optimal value for the parameter $\tau$, balancing detection accuracy and minimizing false positives and negatives. The fifth set of investigations comprehensively evaluated the proposed model's performance, considering accuracy, segmentation quality, computational efficiency, and robustness in challenging scenarios.
The initial ablation experiments aimed to find the optimal hyperparameters $\beta_{1,2}$ in the $L_t$ loss function for achieving the best segmentation performance across different datasets. Varying $\beta_1$ from 0.1 to 0.9 and calculating $\beta_2=1-\beta_1$, the model was trained and evaluated using different combinations of these values. The results demonstrated that assigning a higher weight to $\beta_1$, particularly 0.9, led to superior performance. For example, with $\beta_1=0.9$ and $\beta_2=0.1$, the model achieved high mAP scores on the KUTomaData, Laboro Tomato, and Rob2Pheno Annotated Tomato datasets. Based on these findings, the combination of $\beta_1=0.9$ and $\beta_2=0.2$ was chosen as the optimal hyperparameter choice for subsequent model training, resulting in favourable performance. \newline
Various pre-trained models were integrated into the proposed framework for tomato object segmentation and detection in the ablation experiments and backbone analysis. The performance of these models was compared against the proposed backbone, designed explicitly for this task. The results, summarized in Table \ref{tab:backbone}, clearly demonstrate the superiority of the proposed encoder backbone. Compared to state-of-the-art models such as HRNet, Lite-HRNet, EfficientNet-B4, DenseNet-201, and ResNet-101, the proposed backbone achieved notable improvements across different evaluation metrics. On the KUTomaData dataset, it outperformed existing models by 3.22\%, 2.51\%, 3.67\%, and 0.56\% in terms of $\mu$IoU, $\mu$DC, mAP, and AUC scores, respectively. Similar performance gains were observed on the Laboro and Rob2Pheno datasets, with improvements ranging from 1.60\% to 3.68\% in various evaluation metrics. These significant improvements can be attributed to integrating a novel butterfly structure in the encoder backbone, incorporating distinctive SPB, IB, and HDB blocks. This integration enables the model to extract unique latent characteristics from input images, improving performance in tomato object segmentation and classification tasks. Despite the higher computational cost compared to Lite-HRNet, the selection of the proposed scheme was justified by its superior detection performance. The primary objective of achieving the highest possible detection performance drove this decision. Integrating the butterfly structure and distinctive blocks enables the model to capture essential features and accurate tomato object delineation.\newline
The proposed $L_t$ loss function incorporates a temperature constant ($\tau$) as a hyperparameter to soften target probabilities, improving tomato object segmentation and detection. Adjusting $\tau$ makes the model more receptive to recognizing tomato objects, independent of input imagery characteristics. This softening effect allows the model to comprehend target probabilities better, resulting in enhanced performance. Varying $\tau$ from 1 to 2.5 during training, experiments revealed that increasing $\tau$ from 1 to 1.5 led to significant performance improvements across datasets. For instance, on the KUTomaData dataset, improvements of 4.12\% in $\mu$IoU, 3.21\% in $\mu$DC, 1.65\% in mAP, and 1.25\% in AUC scores were achieved. However, performance declined when $\tau$ exceeded 1.5, indicating a reduced ability to differentiate between object categories. Based on optimal results with $\tau=1.5$, subsequent experiments used this value to balance model receptiveness and accurate classification and segmentation of tomato objects.
In the fifth set of experiments, conducted with the optimal loss function, the performance of the proposed model trained with the $L_t$ loss function was compared against other state-of-the-art loss functions, including $L_{sn}$, $L_{ft}$, $L_{de}$, and $L_{ce}$. The results, summarized in Table \ref{tab:lossFunction}, clearly demonstrated the superiority of the proposed model trained with the $L_t$ loss function across all datasets.
On the KUTomaData dataset, the $L_t$ loss function achieved improvements of 2.16\% in $\mu$IoU, 1.66\% in $\mu$DC, 2.39\% in mAP, and 2.25\% in AUC scores compared to other loss functions. Similarly, on the Laboro dataset, the $L_t$ loss function outperformed the alternatives, resulting in enhancements of 3.25\% in $\mu$IoU, 2.30\% in $\mu$DC, 5.58\% in mAP, and 4.81\% in AUC scores. Furthermore, on the Rob2Pheno Annotated dataset, the $L_t$ loss function delivered improvements of 1.23\% in $\mu$IoU, 0.82\% in $\mu$DC, 1.16\% in mAP, and 1.45\% in AUC scores.
Overall, the proposed framework demonstrates promising results in segmenting and grading tomatoes based on their maturity levels. The experimental analysis validates the effectiveness of the proposed method and highlights its superiority over existing approaches. The framework's robustness to various challenging scenarios and its computational efficiency makes it a valuable tool for assessing tomato quality in greenhouse farming.

\section{Limitations}
\label{l}

In this section, the authors discuss the limitations of the proposed framework and our dataset, along with potential solutions to mitigate them.

\subsection{Limitations of the proposed framework:}

The first limitation of the framework is its inability to generate small masks for extremely occluded, cluttered, or rarely observed small-sized tomatoes.
To address this limitation, a practical approach is to incorporate morphological opening operations as a post-processing step to enhance the quality of small masks. This technique could improve the framework's performance in segmenting such challenging instances.
 
The second limitation of the proposed framework lies in its generation of false masks for highly complex and occluded tomato objects.
Although the produced masks are of decent quality and outperform state-of-the-art methods (as demonstrated in Figure \ref{fig:results_cluttered}), this limitation can still be mitigated by employing more sophisticated segmentation loss functions, such as dice or IoU loss, as objective functions. By utilizing these functions, the model can be constrained to preserve the exact shape of segmented objects, thus reducing the generation of false masks.

Finally, the third limitation of the proposed framework is its potential to generate pixel-level false positives.
This limitation can be overcome by incorporating morphological blob opening operations as a post-processing step, which can effectively eliminate small false positives and improve the overall accuracy of the framework.

In conclusion, While the proposed framework has certain limitations, they can be addressed by integrating appropriate post-processing steps and using more advanced segmentation loss functions during training.
Considering these solutions, the framework can enhance its ability to segment occluded, cluttered, accurately, and rarely observed objects, establishing itself as a more robust solution for tomato object detection.

\subsection{Limitations of the proposed dataset:}

The proposed dataset has the following limitations. Firstly, the tomato dataset may exhibit limited diversity regarding varieties, growth stages, and lighting conditions.
This narrow scope of variation poses a potential drawback, as it may result in overfitting the model to the specific characteristics of the dataset. Consequently, the model's ability to generalize to different scenarios could be compromised.

Secondly, the tomato dataset may contain minor annotation errors, such as inaccurate masking of tomatoes or mislabeling of instances.
These errors can affect the model's performance, making it challenging to achieve high accuracy. To mitigate this limitation, it is essential to thoroughly evaluate and validate all labelled data before utilizing it for training the proposed model.

Lastly, the proposed dataset may primarily cover a specific domain, such as a greenhouse, and may not be suitable for applications in other open-field testing scenarios.
This limited domain coverage can restrict the applicability of models trained solely on this dataset. To address this limitation, it is advisable to incorporate open-field data during training to ensure the models are more adaptable to diverse environments.

By acknowledging and addressing these limitations, the authors can enhance the quality and applicability of the dataset, ultimately facilitating the development of more robust and versatile models for tomato object detection and segmentation.

\section{Conclusions}
\label{c}

This study introduces a novel convolutional transformer-based segmentation and a new dataset of tomato images obtained from greenhouse farms in Al Ajban, Abu Dhabi, UAE. The KUTomaData dataset encompasses images captured under different environmental conditions, including varying light conditions, weather patterns, and stages of plant growth. These factors introduce complexity and challenges for segmentation models in accurately identifying and distinguishing different components of tomato plants. The availability of such a dataset is crucial for developing more precise segmentation models in the robotic harvesting industry, aiming to enhance field efficiency and productivity.
the authors qualitatively assessed and compared our proposed architecture with SETR \cite{setr}, SegFormer \cite{xie2021segformer}, DeepFruits \cite{deepfruits}, COS \cite{COS}, CWD \cite{CWD} and DLIS \cite{Horticulture}. The results demonstrate the superiority of the proposed model across all metrics. 
It outperformed in terms of $\mu$IoU, $\mu$DC, mAP, and AUC across the KUTomaData, Laboro and Rob2Pheno datasets. The results are presented in Table \ref{tab:quan}. 
Moreover, the proposed model exhibits higher class-wise IoU scores for all three tomato ripeness classes, indicating its effectiveness in accurately segmenting each class. 
This work contributes substantially to the computer vision and machine learning community by providing a new dataset that facilitates developing and testing segmentation models specifically designed for agricultural purposes.
Furthermore, it emphasizes the importance of ongoing research and progress in precision agriculture.
In conclusion, the proposed framework and the accompanying KUTomaData dataset contribute to tomato recognition and maturity level classification. The framework addresses the challenges associated with tomato harvesting in real-world scenarios, while the dataset provides a dedicated resource for training and benchmarking deep learning models. The exceptional performance demonstrated by the proposed framework across multiple datasets validates its effectiveness and superiority over existing approaches. Future research can focus on further enhancing the framework's capabilities and exploring its applicability in other agricultural domains.


\section*{Acknowledgements}
This research is supported by ASPIRE, the technology program management pillar of Abu Dhabi’s Advanced Technology Research Council (ATRC), under the ASPIRE project “Aspire Research Institute for Food Security in the Drylands '' within Theme 1.4.

\section*{Data availability}
The data that support the findings of this study are available from ASPIRE, Abu Dhabi, but restrictions apply to the availability of these data, which were used under license for the current study and so are not publicly available. Data are, however, available from the authors upon reasonable request and with permission of ASPIRE, Abu Dhabi.

\section*{Clarification}
1. Please clarify if humans were involved in the study or not?
   \textbf{ No, This study does not involve any human}. \\
2. Experimental research and field studies on plants (either cultivated or wild), including the collection of plant material, must comply with relevant institutional, national, and international guidelines and legislation. We recommend that authors comply with the IUCN Policy Statement on Research Involving Species at Risk of Extinction and the Convention on the Trade in Endangered Species of Wild Fauna and Flora. Please include a statement in this regard.
[ Please clarify if plants were directly used in the study or not ]
\textbf{In this study, plants were not directly used or cultivated.}





\end{document}